\documentclass[letterpaper, 10 pt, conference]{IEEEtran}
\IEEEoverridecommandlockouts

\usepackage[T1]{fontenc}
\usepackage{amsmath,amsfonts} 
\usepackage{amssymb, amsthm}
\usepackage{graphicx}

\usepackage{geometry}
\usepackage{afterpage}

\makeatletter
\let\MYcaption\@makecaption
\makeatother
\usepackage[font=footnotesize]{subcaption}
\makeatletter
\let\@makecaption\MYcaption
\makeatother

\usepackage{booktabs}
\usepackage{color}
\usepackage{microtype}
\usepackage{balance}
\usepackage{cite}

\usepackage{amssymb}
\usepackage{amsfonts}
\usepackage{mathrsfs}
\usepackage{xspace}
\usepackage{bm}
\usepackage{upgreek}

\newcommand{\safemath}[2]{\newcommand{#1}{\ensuremath{#2}\xspace}}

\safemath{\bma}{\mathbf{a}}
\safemath{\bmb}{\mathbf{b}}
\safemath{\bmc}{\mathbf{c}}
\safemath{\bmd}{\mathbf{d}}
\safemath{\bme}{\mathbf{e}}
\safemath{\bmf}{\mathbf{f}}
\safemath{\bmg}{\mathbf{g}}
\safemath{\bmh}{\mathbf{h}}
\safemath{\bmi}{\mathbf{i}}
\safemath{\bmj}{\mathbf{j}}
\safemath{\bmk}{\mathbf{k}}
\safemath{\bml}{\mathbf{l}}
\safemath{\bmm}{\mathbf{m}}
\safemath{\bmn}{\mathbf{n}}
\safemath{\bmo}{\mathbf{o}}
\safemath{\bmp}{\mathbf{p}}
\safemath{\bmq}{\mathbf{q}}
\safemath{\bmr}{\mathbf{r}}
\safemath{\bms}{\mathbf{s}}
\safemath{\bmt}{\mathbf{t}}
\safemath{\bmu}{\mathbf{u}}
\safemath{\bmv}{\mathbf{v}}
\safemath{\bmw}{\mathbf{w}}
\safemath{\bmx}{\mathbf{x}}
\safemath{\bmy}{\mathbf{y}}
\safemath{\bmz}{\mathbf{z}}
\safemath{\bmzero}{\mathbf{0}}
\safemath{\bmone}{\mathbf{1}}

\bmdefine{\biad}{a}
\bmdefine{\bibd}{b}
\bmdefine{\bicd}{c}
\bmdefine{\bidd}{d}
\bmdefine{\bied}{e}
\bmdefine{\bifd}{f}
\bmdefine{\bigd}{g}
\bmdefine{\bihd}{h}
\bmdefine{\biid}{i}
\bmdefine{\bijd}{j}
\bmdefine{\bikd}{k}
\bmdefine{\bild}{l}
\bmdefine{\bimd}{m}
\bmdefine{\bind}{n}
\bmdefine{\biod}{o}
\bmdefine{\bipd}{p}
\bmdefine{\biqd}{q}
\bmdefine{\bird}{r}
\bmdefine{\bisd}{s}
\bmdefine{\bitd}{t}
\bmdefine{\biud}{u}
\bmdefine{\bivd}{v}
\bmdefine{\biwd}{w}
\bmdefine{\bixd}{x}
\bmdefine{\biyd}{y}
\bmdefine{\bizd}{z}

\bmdefine{\bixid}{\xi}
\bmdefine{\bilambdad}{\lambda}
\bmdefine{\bimud}{\mu}
\bmdefine{\bithetad}{\theta}
\bmdefine{\biphid}{\phi}
\bmdefine{\bideltad}{\delta}

\safemath{\bmia}{\biad}
\safemath{\bmib}{\bibd}
\safemath{\bmic}{\bicd}
\safemath{\bmid}{\bidd}
\safemath{\bmie}{\bied}
\safemath{\bmif}{\bifd}
\safemath{\bmig}{\bigd}
\safemath{\bmih}{\bihd}
\safemath{\bmii}{\biid}
\safemath{\bmij}{\bijd}
\safemath{\bmik}{\bikd}
\safemath{\bmil}{\bild}
\safemath{\bmim}{\bimd}
\safemath{\bmin}{\bind}
\safemath{\bmio}{\biod}
\safemath{\bmip}{\bipd}
\safemath{\bmiq}{\biqd}
\safemath{\bmir}{\bird}
\safemath{\bmis}{\bisd}
\safemath{\bmit}{\bitd}
\safemath{\bmiu}{\biud}
\safemath{\bmiv}{\bivd}
\safemath{\bmiw}{\biwd}
\safemath{\bmix}{\bixd}
\safemath{\bmiy}{\biyd}
\safemath{\bmiz}{\bizd}

\safemath{\bmxi}{\bixid}
\safemath{\bmlambda}{\bilambdad}
\safemath{\bmmu}{\bimud}
\safemath{\bmtheta}{\bithetad}
\safemath{\bmphi}{\biphid}
\safemath{\bmdelta}{\bideltad}

\safemath{\bA}{\mathbf{A}}
\safemath{\bB}{\mathbf{B}}
\safemath{\bC}{\mathbf{C}}
\safemath{\bD}{\mathbf{D}}
\safemath{\bE}{\mathbf{E}}
\safemath{\bF}{\mathbf{F}}
\safemath{\bG}{\mathbf{G}}
\safemath{\bH}{\mathbf{H}}
\safemath{\bI}{\mathbf{I}}
\safemath{\bJ}{\mathbf{J}}
\safemath{\bK}{\mathbf{K}}
\safemath{\bL}{\mathbf{L}}
\safemath{\bM}{\mathbf{M}}
\safemath{\bN}{\mathbf{N}}
\safemath{\bO}{\mathbf{O}}
\safemath{\bP}{\mathbf{P}}
\safemath{\bQ}{\mathbf{Q}}
\safemath{\bR}{\mathbf{R}}
\safemath{\bS}{\mathbf{S}}
\safemath{\bT}{\mathbf{T}}
\safemath{\bU}{\mathbf{U}}
\safemath{\bV}{\mathbf{V}}
\safemath{\bW}{\mathbf{W}}
\safemath{\bX}{\mathbf{X}}
\safemath{\bY}{\mathbf{Y}}
\safemath{\bZ}{\mathbf{Z}}

\safemath{\bZero}{\mathbf{0}}
\safemath{\bOne}{\mathbf{1}}
\safemath{\bDelta}{\mathbf{\Delta}}
\safemath{\bLambda}{\mathbf{\UpLambda}}
\safemath{\bPhi}{\mathbf{\Upphi}}
\safemath{\bSigma}{\mathbf{\Upsigma}}
\safemath{\bOmega}{\mathbf{\Upomega}}
\safemath{\bTheta}{\mathbf{\Uptheta}}

\bmdefine{\biAd}{A}
\bmdefine{\biBd}{B}
\bmdefine{\biCd}{C}
\bmdefine{\biDd}{D}
\bmdefine{\biEd}{E}
\bmdefine{\biFd}{F}
\bmdefine{\biGd}{G}
\bmdefine{\biHd}{H}
\bmdefine{\biId}{I}
\bmdefine{\biJd}{J}
\bmdefine{\biKd}{K}
\bmdefine{\biLd}{L}
\bmdefine{\biMd}{M}
\bmdefine{\biOd}{N}
\bmdefine{\biPd}{O}
\bmdefine{\biQd}{P}
\bmdefine{\biRd}{R}
\bmdefine{\biSd}{S}
\bmdefine{\biTd}{T}
\bmdefine{\biUd}{U}
\bmdefine{\biVd}{V}
\bmdefine{\biWd}{W}
\bmdefine{\biXd}{X}
\bmdefine{\biYd}{Y}
\bmdefine{\biZd}{Z}

\bmdefine{\biDelta}{\Delta}
\bmdefine{\biLambda}{\Lambda}
\bmdefine{\biPhi}{\Phi}
\bmdefine{\biSigma}{\Sigma}
\bmdefine{\biOmega}{\Omega}
\bmdefine{\biTheta}{\Theta}

\safemath{\bimA}{\biAd}
\safemath{\bimB}{\biBd}
\safemath{\bimC}{\biCd}
\safemath{\bimD}{\biDd}
\safemath{\bimE}{\biEd}
\safemath{\bimF}{\biFd}
\safemath{\bimG}{\biGd}
\safemath{\bimH}{\biHd}
\safemath{\bimI}{\biId}
\safemath{\bimJ}{\biJd}
\safemath{\bimK}{\biKd}
\safemath{\bimL}{\biLd}
\safemath{\bimM}{\biMd}
\safemath{\bimN}{\biNd}
\safemath{\bimO}{\biOd}
\safemath{\bimP}{\biPd}
\safemath{\bimQ}{\biQd}
\safemath{\bimR}{\biRd}
\safemath{\bimS}{\biSd}
\safemath{\bimT}{\biTd}
\safemath{\bimU}{\biUd}
\safemath{\bimV}{\biVd}
\safemath{\bimW}{\biWd}
\safemath{\bimX}{\biXd}
\safemath{\bimY}{\biYd}
\safemath{\bimZ}{\biZd}

\safemath{\bimDelta}{\biDelta}
\safemath{\bimLambda}{\biLambda}
\safemath{\bimPhi}{\biPhi}
\safemath{\bimSigma}{\biSigma}
\safemath{\bimOmega}{\biOmega}
\safemath{\bimTheta}{\biTheta}

\safemath{\setA}{\mathcal{A}}
\safemath{\setB}{\mathcal{B}}
\safemath{\setC}{\mathcal{C}}
\safemath{\setD}{\mathcal{D}}
\safemath{\setE}{\mathcal{E}}
\safemath{\setF}{\mathcal{F}}
\safemath{\setG}{\mathcal{G}}
\safemath{\setH}{\mathcal{H}}
\safemath{\setI}{\mathcal{I}}
\safemath{\setJ}{\mathcal{J}}
\safemath{\setK}{\mathcal{K}}
\safemath{\setL}{\mathcal{L}}
\safemath{\setM}{\mathcal{M}}
\safemath{\setN}{\mathcal{N}}
\safemath{\setO}{\mathcal{O}}
\safemath{\setP}{\mathcal{P}}
\safemath{\setQ}{\mathcal{Q}}
\safemath{\setR}{\mathcal{R}}
\safemath{\setS}{\mathcal{S}}
\safemath{\setT}{\mathcal{T}}
\safemath{\setU}{\mathcal{U}}
\safemath{\setV}{\mathcal{V}}
\safemath{\setW}{\mathcal{W}}
\safemath{\setX}{\mathcal{X}}
\safemath{\setY}{\mathcal{Y}}
\safemath{\setZ}{\mathcal{Z}}
\safemath{\emptySet}{\varnothing}

\safemath{\colA}{\mathscr{A}}
\safemath{\colB}{\mathscr{B}}
\safemath{\colC}{\mathscr{C}}
\safemath{\colD}{\mathscr{D}}
\safemath{\colE}{\mathscr{E}}
\safemath{\colF}{\mathscr{F}}
\safemath{\colG}{\mathscr{G}}
\safemath{\colH}{\mathscr{H}}
\safemath{\colI}{\mathscr{I}}
\safemath{\colJ}{\mathscr{J}}
\safemath{\colK}{\mathscr{K}}
\safemath{\colL}{\mathscr{L}}
\safemath{\colM}{\mathscr{M}}
\safemath{\colN}{\mathscr{N}}
\safemath{\colO}{\mathscr{O}}
\safemath{\colP}{\mathscr{P}}
\safemath{\colQ}{\mathscr{Q}}
\safemath{\colR}{\mathscr{R}}
\safemath{\colS}{\mathscr{S}}
\safemath{\colT}{\mathscr{T}}
\safemath{\colU}{\mathscr{U}}
\safemath{\colV}{\mathscr{V}}
\safemath{\colW}{\mathscr{W}}
\safemath{\colX}{\mathscr{X}}
\safemath{\colY}{\mathscr{Y}}
\safemath{\colZ}{\mathscr{Z}}

\safemath{\opA}{\mathbb{A}}
\safemath{\opB}{\mathbb{B}}
\safemath{\opC}{\mathbb{C}}
\safemath{\opD}{\mathbb{D}}
\safemath{\opE}{\mathbb{E}}
\safemath{\opF}{\mathbb{F}}
\safemath{\opG}{\mathbb{G}}
\safemath{\opH}{\mathbb{H}}
\safemath{\opI}{\mathbb{I}}
\safemath{\opJ}{\mathbb{J}}
\safemath{\opK}{\mathbb{K}}
\safemath{\opL}{\mathbb{L}}
\safemath{\opM}{\mathbb{M}}
\safemath{\opN}{\mathbb{N}}
\safemath{\opO}{\mathbb{O}}
\safemath{\opP}{\mathbb{P}}
\safemath{\opQ}{\mathbb{Q}}
\safemath{\opR}{\mathbb{R}}
\safemath{\opS}{\mathbb{S}}
\safemath{\opT}{\mathbb{T}}
\safemath{\opU}{\mathbb{U}}
\safemath{\opV}{\mathbb{V}}
\safemath{\opW}{\mathbb{W}}
\safemath{\opX}{\mathbb{X}}
\safemath{\opY}{\mathbb{Y}}
\safemath{\opZ}{\mathbb{Z}}
\safemath{\opZero}{\mathbb{O}}
\safemath{\identityop}{\opI}

\safemath{\veca}{\bma}
\safemath{\vecb}{\bmb}
\safemath{\vecc}{\bmc}
\safemath{\vecd}{\bmd}
\safemath{\vece}{\bme}
\safemath{\vecf}{\bmf}
\safemath{\vecg}{\bmg}
\safemath{\vech}{\bmh}
\safemath{\veci}{\bmi}
\safemath{\vecj}{\bmj}
\safemath{\veck}{\bmk}
\safemath{\vecl}{\bml}
\safemath{\vecm}{\bmm}
\safemath{\vecn}{\bmn}
\safemath{\veco}{\bmo}
\safemath{\vecp}{\bmp}
\safemath{\vecq}{\bmq}
\safemath{\vecr}{\bmr}
\safemath{\vecs}{\bms}
\safemath{\vect}{\bmt}
\safemath{\vecu}{\bmu}
\safemath{\vecv}{\bmv}
\safemath{\vecw}{\bmw}
\safemath{\vecx}{\bmx}
\safemath{\vecy}{\bmy}
\safemath{\vecz}{\bmz}

\safemath{\veczero}{\bmzero}
\safemath{\vecone}{\bmone}
\safemath{\vecxi}{\bmxi}
\safemath{\veclambda}{\bmlambda}
\safemath{\vecmu}{\bmmu}
\safemath{\vectheta}{\bmtheta}
\safemath{\vecphi}{\bmphi}
\safemath{\vecdelta}{\bmdelta}

\safemath{\matA}{\bA}
\safemath{\matB}{\bB}
\safemath{\matC}{\bC}
\safemath{\matD}{\bD}
\safemath{\matE}{\bE}
\safemath{\matF}{\bF}
\safemath{\matG}{\bG}
\safemath{\matH}{\bH}
\safemath{\matI}{\bI}
\safemath{\matJ}{\bJ}
\safemath{\matK}{\bK}
\safemath{\matL}{\bL}
\safemath{\matM}{\bM}
\safemath{\matN}{\bN}
\safemath{\matO}{\bO}
\safemath{\matP}{\bP}
\safemath{\matQ}{\bQ}
\safemath{\matR}{\bR}
\safemath{\matS}{\bS}
\safemath{\matT}{\bT}
\safemath{\matU}{\bU}
\safemath{\matV}{\bV}
\safemath{\matW}{\bW}
\safemath{\matX}{\bX}
\safemath{\matY}{\bY}
\safemath{\matZ}{\bZ}
\safemath{\matzero}{\bmzero}

\safemath{\matDelta}{\bDelta}
\safemath{\matLambda}{\bLambda}
\safemath{\matPhi}{\bPhi}
\safemath{\matSigma}{\bSigma}
\safemath{\matOmega}{\bOmega}
\safemath{\matTheta}{\bTheta}

\safemath{\matidentity}{\matI}
\safemath{\matone}{\matO}

\safemath{\rnda}{A}
\safemath{\rndb}{B}
\safemath{\rndc}{C}
\safemath{\rndd}{D}
\safemath{\rnde}{E}
\safemath{\rndf}{F}
\safemath{\rndg}{G}
\safemath{\rndh}{H}
\safemath{\rndi}{I}
\safemath{\rndj}{J}
\safemath{\rndk}{K}
\safemath{\rndl}{L}
\safemath{\rndm}{M}
\safemath{\rndn}{N}
\safemath{\rndo}{O}
\safemath{\rndp}{P}
\safemath{\rndq}{Q}
\safemath{\rndr}{R}
\safemath{\rnds}{S}
\safemath{\rndt}{T}
\safemath{\rndu}{U}
\safemath{\rndv}{V}
\safemath{\rndw}{W}
\safemath{\rndx}{X}
\safemath{\rndy}{Y}
\safemath{\rndz}{Z}

\safemath{\rveca}{\bimA}
\safemath{\rvecb}{\bimB}
\safemath{\rvecc}{\bimC}
\safemath{\rvecd}{\bimD}
\safemath{\rvece}{\bimE}
\safemath{\rvecf}{\bimF}
\safemath{\rvecg}{\bimG}
\safemath{\rvech}{\bimH}
\safemath{\rveci}{\bimI}
\safemath{\rvecj}{\bimJ}
\safemath{\rveck}{\bimK}
\safemath{\rvecl}{\bimL}
\safemath{\rvecm}{\bimM}
\safemath{\rvecn}{\bimN}
\safemath{\rveco}{\bomO}
\safemath{\rvecp}{\bimP}
\safemath{\rvecq}{\bimQ}
\safemath{\rvecr}{\bimR}
\safemath{\rvecs}{\bimS}
\safemath{\rvect}{\bimT}
\safemath{\rvecu}{\bimU}
\safemath{\rvecv}{\bimV}
\safemath{\rvecw}{\bimW}
\safemath{\rvecx}{\bimX}
\safemath{\rvecy}{\bimY}
\safemath{\rvecz}{\bimZ}

\safemath{\rvecxi}{\bmxi}
\safemath{\rveclambda}{\bmlambda}
\safemath{\rvecmu}{\bmmu}
\safemath{\rvectheta}{\bmtheta}
\safemath{\rvecphi}{\bmphi}

\safemath{\rmatA}{\bimA}
\safemath{\rmatB}{\bimB}
\safemath{\rmatC}{\bimC}
\safemath{\rmatD}{\bimD}
\safemath{\rmatE}{\bimE}
\safemath{\rmatF}{\bimF}
\safemath{\rmatG}{\bimG}
\safemath{\rmatH}{\bimH}
\safemath{\rmatI}{\bimI}
\safemath{\rmatJ}{\bimJ}
\safemath{\rmatK}{\bimK}
\safemath{\rmatL}{\bimL}
\safemath{\rmatM}{\bimM}
\safemath{\rmatN}{\bimN}
\safemath{\rmatO}{\bimO}
\safemath{\rmatP}{\bimP}
\safemath{\rmatQ}{\bimQ}
\safemath{\rmatR}{\bimR}
\safemath{\rmatS}{\bimS}
\safemath{\rmatT}{\bimT}
\safemath{\rmatU}{\bimU}
\safemath{\rmatV}{\bimV}
\safemath{\rmatW}{\bimW}
\safemath{\rmatX}{\bimX}
\safemath{\rmatY}{\bimY}
\safemath{\rmatZ}{\bimZ}

\safemath{\rmatDelta}{\bimDelta}
\safemath{\rmatLambda}{\bimLambda}
\safemath{\rmatPhi}{\bimPhi}
\safemath{\rmatSigma}{\bimSigma}
\safemath{\rmatOmega}{\bimOmega}
\safemath{\rmatTheta}{\bimTheta}

\usepackage{amssymb}
\usepackage{amsfonts}
\usepackage{mathrsfs}
\usepackage{xspace}
\usepackage{bm}
\usepackage{fancyref}
\usepackage{textcomp}

\usepackage{multirow}
\usepackage{stmaryrd}

\newenvironment{textbmatrix}{	\setlength{\arraycolsep}{2.5pt}%
								\big[\begin{matrix}}{\end{matrix}\big]%
								\raisebox{0.08ex}{\vphantom{M}}}

\def\be{\begin{equation}}
\def\ee{\end{equation}}
\def\een{\nonumber \end{equation}}
\def\mat{\begin{bmatrix}}
\def\emat{\end{bmatrix}}
\def\btm{\begin{textbmatrix}}
\def\etm{\end{textbmatrix}}

\def\ba#1\ea{\begin{align}#1\end{align}}
\def\bas#1\eas{\begin{align*}#1\end{align*}}
\def\bs#1\es{\begin{split}#1\end{split}} 
\def\bg#1\eg{\begin{gather}#1\end{gather}}
\def\bml#1\eml{\begin{multline}#1\end{multline}}
\def\bi#1\ei{\begin{itemize}#1\end{itemize}}

\safemath{\dirac}{\delta}					
\safemath{\krond}{\dirac}					

\safemath{\upto}{\uparrow}
\safemath{\downto}{\downarrow}
\safemath{\iu}{j}							
\safemath{\ev}{\lambda}						
\safemath{\hilseqspace}{l^{2}}				
\newcommand{\banachfunspace}[1]{\setL^{#1}}	
\safemath{\hilfunspace}{\banachfunspace{2}}	

\safemath{\SNR}{\textsf{SNR}} 				
\safemath{\PAR}{\textsf{PAR}} 				
\safemath{\No}{N_0}							
\safemath{\Es}{E_s}							
\safemath{\Eb}{E_b}							
\safemath{\EbNo}{\frac{\Eb}{\No}}
\safemath{\EsNo}{\frac{\Es}{\No}}

\DeclareMathOperator{\CHop}{\ensuremath{\opH}} 
\safemath{\tvir}{\rndh_{\CHop}}				
\safemath{\tvtf}{\rndl_{\CHop}}				
\safemath{\spf}{\rnds_{\CHop}}				
\safemath{\bff}{H_{\CHop}}					

\safemath{\ircf}{r_{h}}						
\safemath{\tftvcf}{r_{s}}					
\safemath{\tfcf}{r_{l}}						
\safemath{\bfcf}{r_{H}}						

\safemath{\tcorr}{c_h}						
\safemath{\scf}{c_{s}}						
\safemath{\tfcorr}{c_{l}}					
\safemath{\fcorr}{c_{H}}						

\safemath{\mi}{I}							
\safemath{\capacity}{C}						

\safemath{\normal}{\mathcal{N}}			
\safemath{\jpg}{\mathcal{CN}}			
\safemath{\mchain}{\leftrightarrow}		

\safemath{\dB}{\,\mathrm{dB}}
\safemath{\dBm}{\,\mathrm{dBm}}
\safemath{\Hz}{\,\mathrm{Hz}}
\safemath{\kHz}{\,\mathrm{kHz}}
\safemath{\MHz}{\,\mathrm{MHz}}
\safemath{\GHz}{\,\mathrm{GHz}}
\safemath{\s}{\,\mathrm{s}}
\safemath{\ms}{\,\mathrm{ms}}
\safemath{\mus}{\,\mathrm{\text{\textmu}s}}
\safemath{\ns}{\,\mathrm{ns}}
\safemath{\ps}{\,\mathrm{ps}}
\safemath{\meter}{\,\mathrm{m}}
\safemath{\mm}{\,\mathrm{mm}}
\safemath{\cm}{\,\mathrm{cm}}
\safemath{\m}{\,\mathrm{m}}
\safemath{\W}{\,\mathrm{W}}
\safemath{\mW}{\, \mathrm{mW}}
\safemath{\J}{\,\mathrm{J}}
\safemath{\K}{\,\mathrm{K}}
\safemath{\bit}{\,\mathrm{bit}}
\safemath{\nat}{\,\mathrm{nat}}

\safemath{\define}{\triangleq}			

\safemath{\equivalent}{\sim}
\safemath{\distas}{\sim}					
\safemath{\sdiff}{\Delta}				

\safemath{\reals}{\mathbb{R}}
\safemath{\positivereals}{\reals_{+}}
\safemath{\integers}{\mathbb{Z}}
\safemath{\posint}{\integers_{+}}
\safemath{\naturals}{\mathbb{N}}
\safemath{\posnaturals}{\naturals_{+}}
\safemath{\complexset}{\mathbb{C}}
\safemath{\rationals}{\mathbb{Q}}

\newcommand*{\fancyrefapplabelprefix}{app}		
\newcommand*{\fancyrefthmlabelprefix}{thm}		
\newcommand*{\fancyreflemlabelprefix}{lem}		
\newcommand*{\fancyrefcorlabelprefix}{cor}		
\newcommand*{\fancyrefdeflabelprefix}{def}		
\newcommand*{\fancyrefproplabelprefix}{prop}	
\newcommand*{\fancyrefobslabelprefix}{obs}		
\newcommand*{\fancyrefalglabelprefix}{alg}		
\newcommand*{\fancyrefasmlabelprefix}{asm}	    
\newcommand*{\fancyrefasmslabelprefix}{asms}	    
\newcommand*{\fancyreftbllabelprefix}{tbl}	    
\newcommand*{\fancyreftremabelprefix}{rem}	    

\frefformat{vario}{\fancyrefseclabelprefix}{Section~#1}
\frefformat{vario}{\fancyrefthmlabelprefix}{Theorem~#1}
\frefformat{vario}{\fancyreflemlabelprefix}{Lemma~#1}
\frefformat{vario}{\fancyrefcorlabelprefix}{Corollary~#1}
\frefformat{vario}{\fancyrefdeflabelprefix}{Definition~#1}
\frefformat{vario}{\fancyrefobslabelprefix}{Observation~#1}
\frefformat{vario}{\fancyrefasmlabelprefix}{Assumption~#1}
\frefformat{vario}{\fancyrefasmslabelprefix}{Assumptions~#1}
\frefformat{vario}{\fancyreffiglabelprefix}{Figure~#1}
\frefformat{vario}{\fancyrefapplabelprefix}{Appendix~#1} 
\frefformat{vario}{\fancyrefproplabelprefix}{Proposition~#1}
\frefformat{vario}{\fancyrefalglabelprefix}{Algorithm~#1}
\frefformat{vario}{\fancyrefeqlabelprefix}{(#1)}
\frefformat{vario}{\fancyreftbllabelprefix}{Table~#1}
\frefformat{vario}{\fancyreftremabelprefix}{Remark~#1}

\safemath{\dictab}{[\,\dicta\,\,\dictb\,]}

\safemath{\ysig}{\bmy}
\safemath{\ysighat}{\hat{\ysig}}
\safemath{\ysigdim}{M}
\safemath{\xsig}{\bmx}
\safemath{\xsigdim}{N}
\safemath{\nx}{n_x}
\safemath{\zsig}{\bmz}
\safemath{\zsigdim}{\ysigdim}
\safemath{\rsig}{\bmr}
\safemath{\Adict}{\bA}
\safemath{\Adicttilde}{\widetilde{\Adict}}
\safemath{\Adictdim}{\outputdim\times\xsigdim}
\safemath{\avec}{\bma}
\safemath{\avectilde}{\tilde{\avec}}
\safemath{\Bdict}{\bB}
\safemath{\Bdicttilde}{\widetilde{\Bdict}}
\safemath{\Cdict}{\bC}
\safemath{\cvec}{\bmc}
\safemath{\Ddict}{\bD}
\safemath{\Ddictdim}{\ysigdim\times\xsigdim}
\safemath{\dvec}{\bmd}
\safemath{\Ddicttilde}{\widetilde{\bD}}
\safemath{\Bonb}{\bB}
\safemath{\bvec}{\bmb}
\safemath{\Bonbdim}{\ysigdim\times\ysigdim}
\safemath{\noise}{\bmn}
\safemath{\noisedim}{\ysigim}
\safemath{\err}{\bme}
\safemath{\errdim}{\ysigdim}
\safemath{\errset}{\setE}
\safemath{\nerr}{n_e}
\safemath{\delop}{\bP_\errset}
\safemath{\delopc}{\bP_{{\errset}^c}}

\safemath{\cplxi}{\imath}
\safemath{\cplxj}{\jmath}

\safemath{\dict}{\matD}
\safemath{\inputdim}{N}		
\safemath{\outputdim}{M}		
\safemath{\sparsity}{S}	
\safemath{\inputdimA}{{N_a}}	
\safemath{\inputdimB}{{N_b}}	
\safemath{\elemA}{{n_a}}	
\safemath{\elemB}{{n_b}}	
\safemath{\resA}{\matR_a}	
\safemath{\resB}{\matR_b}	
\safemath{\subD}{\matS} 
\safemath{\subA}{\matS_a} 
\safemath{\subB}{\matS_b} 
\safemath{\dicta}{\matA} 	
\safemath{\dictb}{\matB} 	
\safemath{\hollowS}{H}
\safemath{\hollowA}{H_a}
\safemath{\hollowB}{H_b}
\safemath{\cross}{Z}
\safemath{\coh}{\mu_d}			
\safemath{\coha}{\mu_a}			
\safemath{\cohb}{\mu_b}			
\safemath{\mubs}{\nu}	
\safemath{\cohm}{\mu_m} 
\safemath{\dictset}{\setD}	
\safemath{\dictsetp}{\dictset(\coh,\coha,\cohb)}	
\safemath{\dictsetgen}{\dictset_\text{gen}}
\safemath{\dictsetgenp}{\dictsetgen(\coh)}
\safemath{\dictsetonb}{\dictset_\text{onb}}
\safemath{\dictsetonbp}{\dictsetonb(\coh)}

\safemath{\leftside}{U}
\safemath{\rightsideA}{R_a}
\safemath{\rightsideB}{R_b}

\safemath{\indexS}{\setI_S} 

\safemath{\na}{n_a}			
\safemath{\nb}{n_b}			
\safemath{\coeffa}{p_i}	
\safemath{\coeffb}{q_j}	
\safemath{\seta}{\setP}		
\safemath{\setb}{\setQ}     
\safemath{\setw}{\setW}	
\safemath{\setz}{\setZ}	
\safemath{\cola}{\veca}		
\safemath{\colb}{\vecb}		
\safemath{\cold}{\vecd}		
\safemath{\inputvec}{\vecx} 	
\safemath{\error}{\vece}	
\safemath{\noiseout}{\vecz} 	
\safemath{\inputvecel}{x}
\safemath{\inputveca}{\vecx_a}
\safemath{\inputvecb}{\vecx_b}
\safemath{\outputvec}{\vecy}	
\safemath{\lambdamin}{\lambda_{\mathrm{min}}}

\safemath{\elltwo}{\ell_2}
\safemath{\ellone}{\ell_1}
\safemath{\ellzero}{\ell_0}
\safemath{\ellinf}{\ell_\infty}
\safemath{\ellinftilde}{\ell_{\widetilde\infty}}
\safemath{\licard}{Z(\coh,\coha,\cohb)}
\safemath{\xsol}{\hat{x}}
\safemath{\xbord}{x_b}		
\safemath{\xstat}{x_s}		
\safemath{\xstatLone}{\tilde{x}_s}
\safemath{\order}{\mathcal{O}} 
\safemath{\scales}{\Theta} 
\safemath{\ones}{\mathbf{1}} 
\safemath{\zeroes}{\mathbf{0}} 
\safemath{\thlone}{\kappa(\coh,\cohb)} 
\safemath{\constoneA}{\delta} 
\safemath{\constoneB}{\epsilon} 
\safemath{\nlarge}{L}				   
\safemath{\sumlarge}{S_\nlarge}
\safemath{\maxlarger}{P_\nlarge}	   
\safemath{\Pzero}{\textrm{P0}}	
\safemath{\Pone}{\textrm{P1}}
\safemath{\vecfir}{\vecw}			 
\safemath{\vecsec}{\vecz}
\safemath{\elvecfir}{w}              
\safemath{\elvecsec}{z}				 
\safemath{\nlargefir}{n}
\safemath{\normout}{\gamma}
\safemath{\auxfun}{h}
\safemath{\supp}{\textrm{supp}}

\safemath{\indexa}{\ell}
\safemath{\indexb}{r}
\safemath{\indexc}{i}
\safemath{\indexd}{j}

\safemath{\project}{P}

\renewcommand{\bml}{\ensuremath{\boldsymbol \ell}}

\newcommand{\eg}[1]{\textcolor{green}{\bf[eg: #1]}}

\title{\vspace{6.5mm}Siamese Neural Networks for  \\ Wireless Positioning and Channel Charting}

\author{\IEEEauthorblockN{Eric Lei$^\text{1}$,  Oscar Casta\~neda$^\text{1}$,  Olav Tirkkonen$^\text{2}$, Tom Goldstein$^\text{3}$, and Christoph Studer$^\text{1}$} \\
	\IEEEauthorblockA{
		\small $^\text{1}$\textit{School of Electrical and Computer Engineering, Cornell University, Ithaca, NY; email:  {studer@cornell.edu}} \\
		$^\text{2}$\textit{School of Electrical Engineering, Aalto University, Finland; e-mail: {olav.tirkkonen@aalto.fi}}\\
		$^\text{3}$\textit{Department of Computer Science, University of Maryland, College Park, MD; e-mail: {tomg@cs.umd.edu}}\thanks{The work of EL, OC, and CS was supported by Xilinx, Inc. and by the US National Science Foundation (NSF) grants  ECCS-1408006, CCF-1535897,  CCF-1652065, CNS-1717559, and ECCS-1824379.  The work of OT was funded in part by the Academy of Finland (grant 319484). The work of TG was supported by the US NSF under grant CCF-1535902 and by the US Office of Naval Research grant \mbox{N00014-17-1-2078}.}\thanks{The authors would like to thank Mengceng He for developing the T-intersection simulator in Unity. The authors  also thank Emre~G\"{o}n\"{u}lta\c{s} and Pengzhi~Huang for their help with representation-constrained autoencoders.}
	}
}

\begin{document}
 
\maketitle

\begin{abstract}
Neural networks have been proposed recently for positioning and channel charting of user equipments (UEs) in wireless systems. Both of these approaches process channel state information (CSI) that is acquired at a multi-antenna base-station in order to learn a function that maps CSI to location information. CSI-based positioning using deep neural networks requires a dataset that contains both CSI and associated location information. Channel charting (CC) only requires CSI information to extract relative position information. Since CC builds on dimensionality reduction, it can be implemented using autoencoders. In this paper, we propose a unified architecture based on Siamese networks that can be used for supervised UE positioning and unsupervised channel charting. In addition, our framework enables semisupervised positioning, where only a small set of location information is available during training. We use simulations to demonstrate that Siamese networks achieve similar or better performance than existing positioning and CC approaches with a single, unified neural network architecture. 
\end{abstract}


\section{Introduction}

Positioning of wireless transmitters under line-of-sight (LoS) propagation  conditions is a  relatively well understood topic~\mbox{\cite{4343996,Gustafsson2005,gu2009survey}}. Corresponding positioning techniques, such as triangulation and trilateration, find widespread use in global navigation satellite systems (GNSSs) and in localization systems that leverage user equipment (UE) access to multiple infrastructure basestations (BSs). 
However, positioning of wireless transmitters is known to be significantly more challenging under  complex propagation conditions~\cite{soubielle2002gps,qi2006time}.
Such conditions appear, for example, in  non-LoS scenarios (which may happen indoors) or for channels with rich multi-path components (which may happen in dense urban areas).
The trend towards communication at high carrier frequencies, such as millimeter-wave systems, further increases the prevalence of complex propagation conditions as channel properties can abruptly change in space~\cite{zhao201328,akdeniz2014millimeter}.

\subsection{Machine-Learning-Based Positioning}

To enable positioning for such challenging propagation conditions, data-driven approaches that combine  channel-state information (CSI) fingerprinting with machine learning methods have been proposed recently \cite{Gao2015,Wang2015,Wang2016,confi,8292280,arnold2018deep,arnold2019novel,pirzadeh2019machine}.
Most of these methods rely on  deep neural networks, which map CSI to position in space. While these approaches have been shown to achieve high accuracy even under challenging propagation conditions, they require large training databases consisting of CSI and true location information. The acquisition of such databases necessitates extensive and repeated measurement campaigns, where CSI and location measurements must be taken at high spatial resolution. 

Channel charting (CC), as proposed in~\cite{cc_paper}, avoids extensive (and expensive) measurement campaigns for applications that do not require absolute positioning capabilities (e.g., for hand-over, cell search, and user grouping).
The principle of CC is to exploit the fact that CSI is high-dimensional, but strongly dependent on UE position, which is low-dimensional. Dimensionality reduction, e.g., by means of Sammon's mapping~\cite{sammon} or autoencoders \cite{HintonSalakhutdinov2006b,van2009dimensionality, bengio, pmlr-v27-baldi12a}, applied to carefully crafted CSI features builds a \emph{channel chart}, in which nearby points correspond to nearby locations in true space.
Unfortunately, absolute (and exact) position information is not available from conventional CC.
To equip CC with absolute localization capabilities,  the paper \cite{huang2019improving} proposed the inclusion of side information into autoencoders. The resulting semisupervised autoencoder is able to include a subset of known spatial locations, without requiring measurements at wavelength scales in space.
Unfortunately, the low-dimensional representations generated by autoencoders do not exhibit any desirable distance properties (e.g., that nearby features should be nearby in the representation space); corresponding distance constraints must be imposed separately~\cite{huang2019improving}.
Furthermore, autoencoders are often difficult to train, and require tedious network and algorithm parameter tuning~\cite{hunter2012selection}. 

\subsection{Contributions}
In this paper, we propose a unified architecture based on Siamese  networks that enables CSI-based localization in supervised,  semisupervised, and unsupervised scenarios, which includes CC. 
The proposed network architecture is a parametric extension of Sammon's mapping \cite{sammon}, which enables the inclusion of side information that results from (i) a (possibly small) set of annotated points in space with known location and (ii) the fact that UEs are moving with finite velocity~\cite{huang2019improving}.
To demonstrate the effectiveness of Siamese networks for CSI-based positioning and CC, we perform simulations for LoS and non-LoS channels, and we compare our approach to conventional neural networks and CC methods that use Sammon's mapping and autoencoders.
We finally use a simulator that models UE movement at a T-intersection to show that a small set of CSI measurements is sufficient to perform accurate positioning with Siamese networks. 

\subsection{Relevant Prior Art}

A number of recent papers have investigated the efficacy of neural networks for CSI-based  positioning \cite{Gao2015,Wang2015,Wang2016,confi,8292280,arnold2018deep,arnold2019novel,pirzadeh2019machine}. All of these methods are supervised and require extensive measurement campaigns to generate large databases consisting of CSI measurements and accurate position information at densely sampled locations in space (often at wavelength scales). 
Channel charting (CC), as proposed in~\cite{cc_paper}, is unsupervised and uses dimensionality reduction to perform relative positioning solely from CSI measurements, without the need of ground-truth position information.
Recent extensions of CC include multi-point CC \cite{deng2018multipoint} for systems with simultaneous connectivity to multiple BSs and semisupervised CC with autoencoders \cite{huang2019improving}, which enables the inclusion of partially-annotated datasets.
The Siamese network proposed in this paper unifies supervised CSI-based neural-network-based positioning with unsupervised CC in a single network architecture.
In addition, our framework is able to match, and even to outperform, existing methods for all these scenarios. 

Siamese networks have been proposed by Bromley \emph{et al.} in 1993 for handwritten signature verification~\cite{OGSiamese}.
Since then, this neural network topology has been used in a broad range of applications, including classification tasks~\cite{oneshot_siamese}, online object tracking \cite{tracking_siamese}, and similarity function learning for images~\cite{patch_siamese}, human faces~\cite{lstm_siamese}, and text \cite{text_siamese}.
The method proposed in this paper combines Siamese networks with Sammon's mapping~\cite{sammon} in order to perform regression for the supervised scenario and parametric dimensionality reduction for CC. Furthermore, the same architecture enables the inclusion of side information that  stems from the measurement process. 


\section{Siamese Neural Networks}
We now discuss the basics of Sammon's mapping and show how a parametric version can be derived using Siamese networks. We then discuss how the resulting architecture can be augmented for supervised and semisupervised learning.

\subsection{Sammon's Mapping Basics}
Sammon's mapping \cite{sammon} takes a dataset $\setX=\{\bmx_n \in \mathbb{R}^D\}_{n=1}^N$ comprising $N$ input vectors with dimension $D$ and maps all vectors to a set $\setY=\{\bmy_n \in \mathbb{R}^{D'}\}_{n=1}^N$ in some $D'$ dimensional space, where typically $D'\ll D$. The goal of Sammon's mapping is to preserve small pairwise distances between the high- and low-dimensional vectors by minimizing the following loss function:
\begin{align} \label{eq:sammonloss}
L(\setY)=  \sum_{n=1}^{N-1}\sum_{m=n+1}^{N}  w_{n,m} \big (\|\bmx_n\!-\!\bmx_m \| \!-\!\|\bmy_n\!-\!\bmy_m \| \big)^2.
\end{align}
The variables of this optimization problem are the low-dimensional vectors in the set $\setY$. 
The parameters $w_{n,m}$ are used to de-weight the importance of pairs of vectors that are dissimilar in high-dimensional space. The common choice for Sammon's mapping is  $w_{n,m}=\|\bmx_n-\bmx_m\|^{-1}$ for $n\neq m$.

As demonstrated in \cite{cc_paper}, Sammon's mapping performs exceptionally well for CC tasks. However,  this technique is nonparametric as  it does not explicitly learn a function $f:\reals^D \to \reals^{D'}$ that maps a new high-dimensional vector $\bmx_{n'}\in\reals^D$ into a corresponding vector $\bmy_{n'}\in\reals^{D'}$ in low-dimensional space. 
For CSI-based positioning and CC, however, we are interested in a function $f$ that maps high-dimensional vectors (that contain CSI features) into low-dimensional vectors (that contain position information). We next describe a neural-network extension to parametric Sammon's mapping. 

\subsection{Siamese Networks for Parametric Sammon's Mapping}
\label{sec:parametricsammon}
Artificial neural networks are well known to be excellent function approximators. It is therefore natural to replace  each low-dimensional vector $\bmy_{n}\in\reals^{D'}$, $n=1,\ldots,N$, in \fref{eq:sammonloss} by the output of a feedforward neural network $\bmy_n=f_\bmtheta(\bmx_n)$ that maps high-dimensional vectors to low-dimensional vectors. The parameters $\bmtheta$ describe the weights and biases of the neural network. 
We simply minimize the loss function
\begin{align} \label{eq:siamesesammon}
& L(\bmtheta) = \notag  \\
&  \sum_{n=1}^{N}\sum_{m=n+1}^{N}  w_{n,m} \big (\|\bmx_n\!-\!\bmx_m \| \!-\!\|f_\bmtheta(\bmx_n)\!-\!f_\bmtheta(\bmx_m) \| \big)^2\!,
\end{align}
which is a parametric version of Sammon's mapping, where $\bmy_n=f_\bmtheta(\bmx_n)$ is the mapping from high to low dimension. 

\begin{figure}[tp]
\centering
\includegraphics[width=0.95\linewidth]{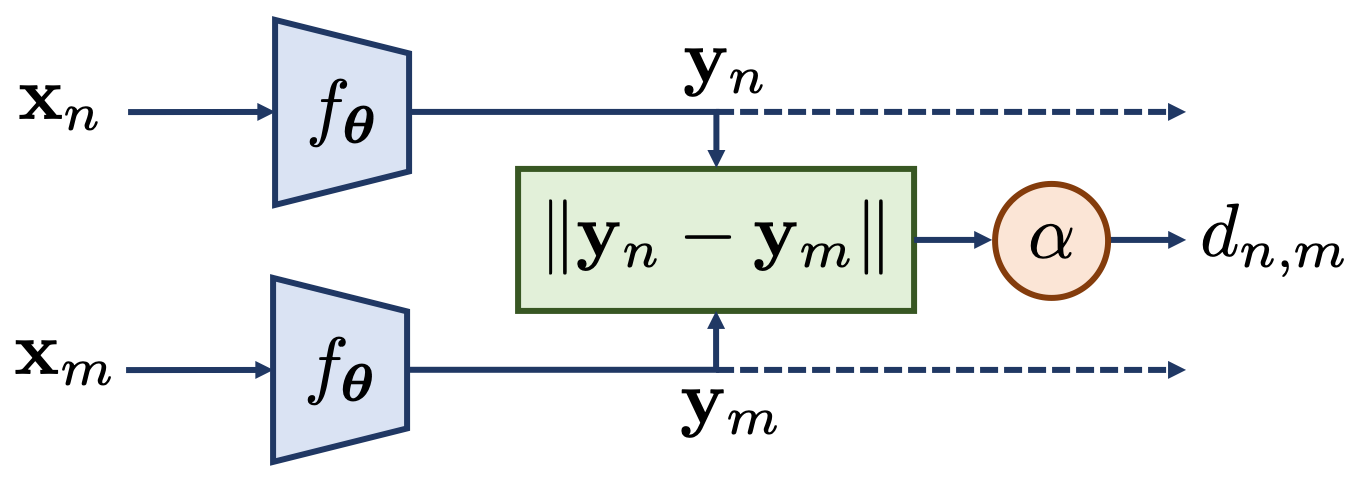}
\vspace{-0.1cm}
\caption{Illustration of the proposed Siamese network. The architecture consists of two parallel neural networks $f_\bmtheta$ that share the same parameters~$\bmtheta$ and process the two high-dimensional vectors $\bmx_n$ and $\bmx_m$. The distance~$d_{n,m}$ between the low-dimensional outputs~$\bmy_n$ and~$\bmy_m$ of the parallel neural networks is the main output of the network. To enable supervised learning, we provide the vectors $\bmy_n$ and $\bmy_m$ as secondary outputs; to enable semisupervised learning, we include a scaling layer that multiplies the output distances of the Siamese network by $\alpha>0$.}
\label{fig:siamese}
\end{figure}

As it turns out, minimizing the loss function in \fref{eq:siamesesammon} is an instance of training a Siamese network~\cite{OGSiamese};  \fref{fig:siamese} illustrates the associated network architecture.
The proposed Siamese network takes in two high-dimensional vectors~$\bmx_n$ and~$\bmx_m$ and maps them to corresponding low-dimensional vectors~$\bmy_n$ and~$\bmy_m$ using two identical neural networks described by the function $f_\bmtheta:\reals^D\to\reals^{D'}$ that share the same set of parameters $\bmtheta$. The scaled distance  between the two low-dimensional vectors  is $d_{n,m} = \alpha \|\bmy_n-\bmy_m\|$, where $\alpha>0$ is a trainable parameter required in semisupervised scenarios (see \fref{sec:supersemisuper}). 

%
Minimizing the loss in \fref{eq:siamesesammon} is equivalent to minimizing a weighted mean-square error (MSE) between the Siamese network's output and the distance $\|\bmx_n-\bmx_m\|$ between pairs of high-dimensional vectors. 
This Siamese network can be trained using the high-dimensional vectors in the set~$\setX$, which makes minimizing \eqref{eq:siamesesammon} an unsupervised learning problem.

	

	
	
\subsection{Supervised and Semisupervised Extensions}
\label{sec:supersemisuper}
In the application of UE positioning, we are often able to acquire ground-truth location information of a subset of the wireless transmitters, e.g., during a dedicated measurement campaign. In the ensuing discussion, we call the case where all $N$ low-dimensional vectors $\bmy_n$ are known a-priori the \emph{supervised} scenario; if only a subset $\setN$ of these vectors is known a-priori, then we call this the \emph{semisupervised} scenario. 

\subsubsection{Supervised scenario}
Siamese networks can easily be extended to support the supervised scenario. 
To include a-priori information on the low-dimensional vectors, we can incorporate additional penalty terms  in the loss function, which depend on the observed low-dimensional vectors~$\bmy_n$.
We include Euclidean distance loss functions between the network's secondary outputs~$\bmy_n$ and~$\bmy_m$, and known low-dimensional vectors. Specifically, we add terms of the form $\|f_\bmtheta(\bmx_n)-\underline{\bmy}_n\|^2$ to the loss function, where underlined quantities, such as $\underline{\bmy}_n$, denote  \emph{known} low-dimensional vectors (also known as anchor points). Note that these terms are equivalent to those of a traditional neural network trained with an MSE loss function.

\subsubsection{Semisupervised scenario}
To support the semisupervised scenario, we include an additional, trainable scaling factor \mbox{$\alpha>0$} prior to computing the distance output $d_{n,m} = \alpha \|\bmy_n-\bmy_m\|$. 
While this scaling factor is irrelevant in the fully supervised and unsupervised scenarios because it can be absorbed in the function $f_\bmtheta$, it is necessary during semisupervised learning for the following reason:
While supervised information is taken into account using the secondary outputs~$\bmy_n$ and~$\bmy_m$ as described above, we also match the Siamese network's distance output $d_{n,m}$ to that of the input distances $\|\bmx_n-\bmx_m\|$. These input-vector distances,  exhibit a different scaling (different to that of $\|\bmy_n-\bmy_m\|$). The learnable factor $\alpha$ enables us to scale the distance output (which is in terms of the low-dimensional vectors) to approximately match distances between high-dimensional vectors $\|\bmx_n-\bmx_m\|$. This trick allows us to train a loss with both an anchor point term and a pair-wise distance term, while having these terms to be consistent.

\subsection{Practical Considerations}
We train the Siamese network in \fref{fig:siamese} using stochastic gradient descent (SGD). To improve the performance of our trained neural networks, we impose an $\ell_2$-norm regularization on all weights. Furthermore, we observed that using large batch sizes improved the quality of the learned networks.
To ensure that (i) the gradient of the loss function is always defined and (ii) the weights $w_{n,m}$ in \fref{eq:siamesesammon} remain bounded, we replace all Euclidean norms in the loss function \fref{eq:siamesesammon}, by the following smooth approximation: $\|\bmx\| \approx \sqrt{\|\bmx\|^2+\varepsilon}.$ Here, $\varepsilon>0$ is a hyper-parameter---typical values for $\varepsilon$ that worked well for our purposes are of the order of $10^{-6}$. 
	
	
\section{Positioning and Channel Charting with Siamese Networks}

We now detail how Siamese networks can be used to perform both  positioning and CC. 

\subsection{Neural-Network-Based Positioning}
Positioning from CSI using neural networks boils down to a simple regression problem \cite{Gao2015,Wang2015,Wang2016,confi,8292280,arnold2018deep,arnold2019novel,pirzadeh2019machine}.
The high-dimensional input vectors $\bmx_n\in\reals^D$ correspond to channel features that are extracted from measured CSI (e.g., obtained during a training phase in the uplink where UEs transmit data to the BS). 
Channel features are usually designed to represent large-scale properties of wireless channels (such as angle-of-arrival, power-delay profile, and receive power), while suppressing noise and small-scale fading artifacts (such as random phase shifts). 
A range of carefully-designed channel features have been proposed in~\cite{cc_paper}.
The low-dimensional vectors  $\bmy_n\in\reals^{D'}$ correspond to measured ground-truth position information, which can be acquired during a dedicated measurement campaign. By forming a large dataset consisting of channel features and associated position vectors, one can simply train a neural network that minimizes the position MSE between ground truth and neural network output in a supervised manner. 

\subsection{Relative Positioning with Channel Charting}
To enable relative positioning without the need of extensive measurement campaigns, CC, as put forward in~\cite{cc_paper}, learns a  low-dimensional \emph{channel chart} that preserves \emph{locally} the original spatial geometry. 
CC collects CSI from a large number of UEs at different spatial locations. The CSI is transformed into channel features,  which are then processed using dimensionality-reduction methods to extract a low-dimensional \emph{channel chart} that preserves \emph{locally} the original spatial geometry. 
UEs that are nearby in real space will be placed nearby in the channel chart and vice versa---global geometry is  not preserved.
CC is unsupervised, meaning that construction of the channel chart is only based on channel features that are collected at a BS.
CC opens up a range of location-based applications without requiring LoS channel conditions, access to GNSS, or extensive measurement campaigns that are needed for fingerprinting  \cite{4343996}.

The operating principle of CC is as follows. Suppose that we have $n=1,\ldots,N$ UE pilot transmissions from coordinates~$\bmy_n^\star\in\reals^{D'}$ where $D'=3$.
The $n$th transmission from location~$\bmy_n^\star$ enables the BS to estimate the high-dimensional CSI vector~$\bmh_n$, which characterizes the channel's multi-path propagation, power delay profile, path loss, etc.
One can now extract channel features $\bmx_n\in\reals^D$ from the CSI vector $\bmh_n$, which only reflect large-scale properties of the wireless channel. 
The key assumption underlying CC is now that large-scale fading characteristics are largely static and are determined by the UE location. 
In fact, due to the underlying physics of wave propagation, each channel feature is a function of the UE position---this function represents the effect of the unknown environment on the transmitted pilot signal. 
From the set of channel features $\{\bmx_n\}_{n=1}^{N}$, one can now learn the channel chart via dimensionality reduction. 

In~\cite{cc_paper,deng2018multipoint,huang2019improving}, Sammon's mapping, Laplacian eigenmaps, and autoencoders have been used to learn  channel charts. 
While Sammon's mapping~\cite{cc_paper} and Laplacian eigenmaps~\cite{deng2018multipoint} have been shown to perform well in LoS and non-LoS scenarios, their nonparametric nature limits their use in practice. 
Very recently, out-of-sample extensions for Laplacian eigenmaps have been proposed in~\cite{ponnada2019out}.
Autoencoders provide a parametric mapping as the encoder function can be used directly for the purpose of mapping  new channel features to (relative) location.
However, autoencoders are often difficult to train~\cite{hunter2012selection} and the inclusion of geometry constraints is not straightforward~\cite{huang2019improving}.
To avoid the drawbacks of existing CC algorithms, we propose to use Siamese networks as illustrated in \fref{fig:siamese} and described in \fref{sec:parametricsammon}.

\subsection{Semi-Supervised Positioning}
In situations where only a subset $\setN\subseteq\{1,2,\ldots,N\}$ of channel features $\bmx_n$, $n\in\setN$, are annotated with ground-truth position information~$\underline{\bmy}_n$, one can combine the ideas of neural-network-based positioning and CC. 
Concretely, we can impose a-priori location information on the secondary outputs $\bmy_n$ and $\bmy_m$ of the Siamese network shown in 
\fref{fig:siamese}. The primary distance output $d_{n,m}$ is used to match pairwise channel feature distances $\|\bmx_n-\bmx_m\|$. 
This approach enables us to \emph{simultaneously} enforce that $\bmy_n=f_\bmtheta(\bmx_n)$ maps to known anchor vectors $\underline{\bmy}_n$, while the scaled pairwise output distances approximately satisfy 
\begin{align}
\alpha\|f_\bmtheta(\bmx_n)-f_\bmtheta(\bmx_m)\| \approx  \|\bmx_n-\bmx_m\|.
\end{align}
During this approach, we learn the neural network parameters~$\bmtheta$ and the scaling factor $\alpha>0$.
We reiterate that the proposed network in \fref{fig:siamese} enables supervised, semisupervised, and unsupervised positioning in a single unified architecture. 
	

%
\section{Results}
We now demonstrate the efficacy of the proposed Siamese network architecture for CSI-based positioning and CC. 
We start by describing the simulated scenario and the used performance metrics.
We then provide a comparison to (i) conventional feedforward neural networks in the supervised scenario and (ii) to Sammon's mapping and autoencoders in the semisupervised scenario and for CC.

\subsection{Simulated Scenario}

\begin{figure}[tp]
\centering
\includegraphics[width=0.75\linewidth]{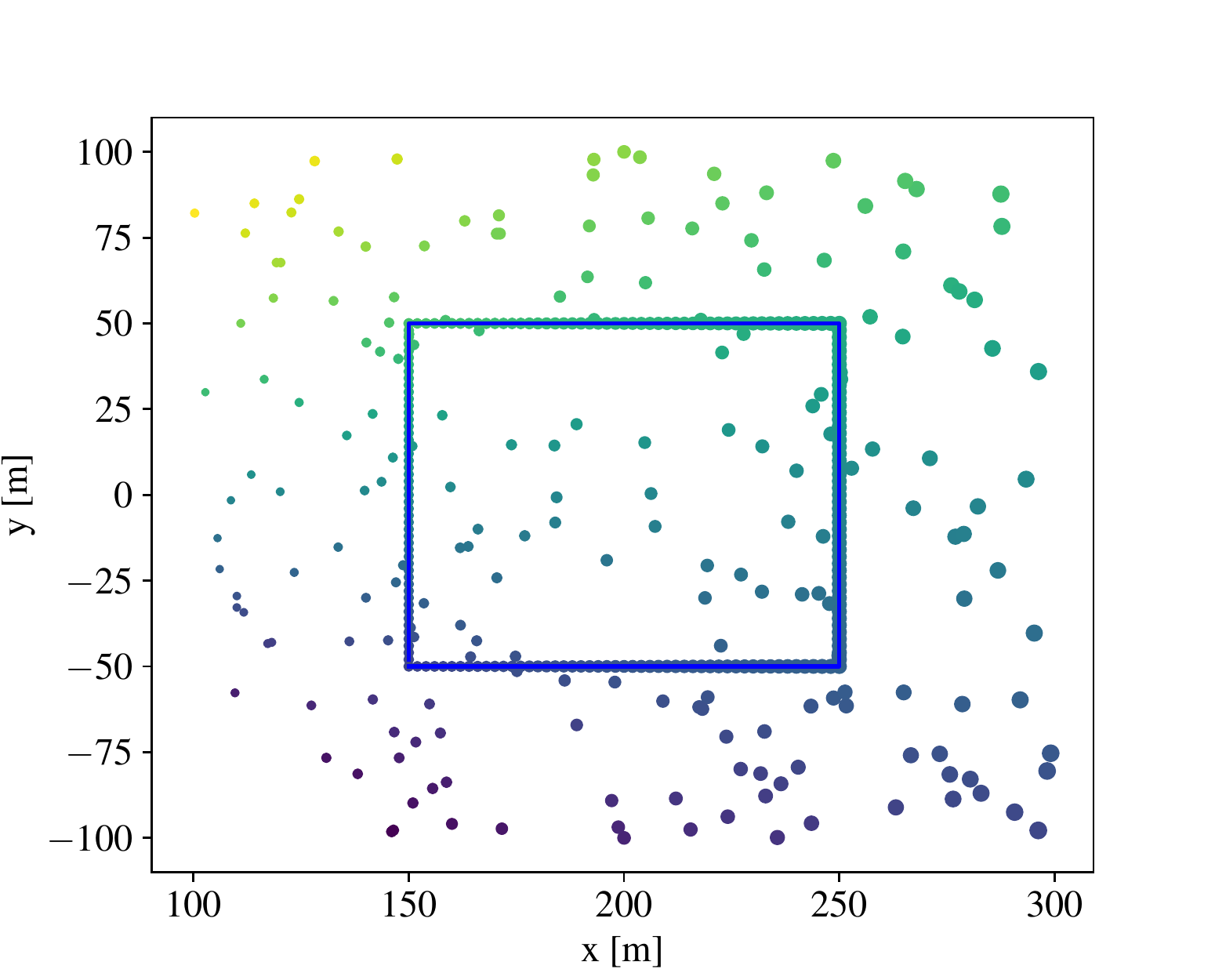}
\caption{Simulated UE locations of the synthetic test dataset. The multi-antenna BS antenna array with $B=32$ antennas is located at the origin $(0,0)$. Each point represents a UE position; 200 UEs are positioned randomly in a rectangular area; 200 UEs are positioned on a square in space to facilitate visualization of supervised and unsupervised positioning methods. The points associated to locations far away from the BS have more thickness; the color gradients further aid visualization of our results.}
\label{fig:scenario} 
\end{figure}

\fref{fig:scenario} depicts the simulated scenario and \fref{tbl:scenario} summarizes the key system and simulation parameters. 
We randomly place $N=2\,000$ UEs in a rectangular area of $200\times200$\,m$^\text{2}$; their associated CSI is used to extract the training data consisting of channel features. We additionally extract a test set composed by $400$ points: $200$ UEs are randomly placed in the same area; $200$ UEs are placed on a square shape to facilitate visualization of the learned position information. 
A  BS with $B=32$ antennas on a uniform linear array (ULA) of $\lambda/2$ spacing is located at  $(x,y,z)=(0,0,30)$ meters. 
The UEs transmit one pilot symbol at $20$\,dBm. 
We model data transmission at $2.68$\,GHz with 20\,MHz bandwidth for two channel models: (i) Quadriga LoS (Q-LoS; a LoS channel model that includes scatterers); and (ii) Quadriga non-LoS (Q-NLoS; a non-LoS channel model which only includes scatterers).
We use the ``Berlin UMa'' scenario for both scenarios~\cite{jaeckel2014quadriga} with spatial consistency enabled.

\begin{table}[tp]
\centering
\renewcommand{\arraystretch}{1.05}
\caption{Summary of system and Quadriga channel model parameters}
\label{tbl:scenario}
\begin{tabular}{@{}ll|ll@{}}
\toprule
Scenario & Berlin UMa & Antenna array & $\lambda/2$ ULA  \\
Carrier frequency & 2.68\,GHz & BS location & (0,0)  \\
Bandwidth & 20\,MHz & BS height & 30\,m  \\
BS antennas & 32 & UE height & 2.5\,m  \\
Subcarriers & 8 & UE transmit power  & $20$\,dBm \\
\bottomrule
\end{tabular}
\end{table}

At the BS side, we extract CSI at $8$ orthogonal-frequency-division multiplexing (OFDM) subcarriers, which results in CSI consisting of $32\times 8$ complex-valued  coefficients (i.e., BS antennas times subcarriers). 
The channel features are computed by first applying feature scaling as in \cite{cc_paper}, and then transforming the scaled features from the antenna and frequency domain to the beamspace and delay domain (by taking a $2$-D discrete Fourier transform across the $32$ antennas and $8$ subcarriers). 
We then take the entry-wise absolute value of the features and stack all $8$ subcarriers to one $256$-dimensional channel-feature vector. The resulting channel feature vectors $\{\bmx_n\}_{n=1}^N$ are of dimension $D=256$; the dimension of the location data is $D'=2$ as we assume the UEs to be at the same height.

For the Siamese network, we use a neural network architecture with $6$ hidden layers, each having the following number of activations per layer: $512$, $256$, $128$, $64$, $32$, and $2$. Layers~$1$ to~$5$ use rectified linear unit (ReLU) activations, whereas the last layer $6$ uses a linear activation.
For the reference fully-connected neural network (FCNN), we use the same network topology. 
For the reference autoencoder (AE) proposed in~\cite{cc_paper,huang2019improving}, we use the same network topology for the encoder and the reverse topology for the decoder network. 

\subsection{Performance Metrics}
To measure the performance of the considered methods, we will use the following metrics:

\subsubsection{Mean distance error}
In the supervised and semisupervised scenarios, a natural way for measuring performance is the mean distance error (MDE), which we define as follows:
\begin{align}
\textit{MDE} = \frac{1}{T}\sum_{t=1}^T \|\bmy_t-\bmy_t^\star\|.
\end{align}
Here, $T$ denotes the number of vectors in the test set, $\bmy_t^\star$ is the low-dimensional ground truth associated with the high-dimensional vector $\bmx_t^\star$, and $\bmy_t=f_\bmtheta(\bmx_t^\star)$ is the output of the learned neural network $f_\bmtheta$. 

\subsubsection{Kruskal's stress}
In addition to the MDE, we use Kruskal's stress (KS)~\cite{lee2009dr, shepard1962} to measure how well the low-dimensional dataset $\{\bmy_n\}_{n=1}^N$ represents the original location dataset $\{\bmy_n^\star\}_{n=1}^N$. Specifically, KS is defined as
\begin{align}
\textit{KS} =  \sqrt{\frac{\sum_{n,m}(\delta_{n,m}-\beta d_{n,m})^2}{\sum_{n,m}{\delta_{n,m}^2}}}.
\end{align}
Here, $\delta_{n,m}=\|\bmy_n^\star-\bmy_m^\star\|$, $d_{n,m}=\|\bmy_n-\bmy_m\|$, and $\beta=\sum_{n,m}\delta_{n,m} d_{n,m}/\sum_{n,m}\delta_{n,m}^2$. KS is in the range $[0,1]$ and smaller values indicate that global geometry is preserved better; if $\textit{KS}=0$, then geometry is perfectly preserved.

\subsubsection{Trustworthiness and continuity}
To characterize the performance of dimensionality reduction, especially in the unsupervised scenario where absolute position information is unavailable, we use two standard metrics that characterize neighborhood preserving properties: \emph{trustworthiness} (TW) and \emph{continuity} (CT)\cite{venna2001neighborhood,kaski2003trustworthiness,vathy2013graph}. 
The TW measures whether the mapping of high-dimensional vectors to the low-dimensional space introduces new (false) neighbors. The TW is defined as 
\begin{align}
\textit{TW}(K) = 1- {\textstyle \frac{2}{NK(2N-3K-1)}}\sum_{n=1}^N \sum_{m\in\setU^K_n}(r(n,m)-K),
\end{align}
where $r(n,m)$ denotes the rank of the low-dimensional vector~$\bmy_n$ among the pairwise distances between the other vectors. The set~$\setU^K_n$ contains the vectors that are among the~$K$ nearest neighbors in low-dimensional space, but not in high-dimensional space. 
The CT measures whether similar vectors in high-dimensional space remain similar in the low-dimensional space, and is defined as  
\begin{align}
\textit{CT}(K) =  1-{\textstyle\frac{2}{NK(2N-3K-1)}}\sum_{n=1}^N \sum_{m\in\setV^K_n} (\hat{r}(n,m)-K),
\end{align}
where $\hat{r}(n,m)$ denotes the rank of the high-dimensional vector~$\bmx_n$ among the pairwise distances between the other vectors. The set $\setV^K_n$ contains the vectors that are among the $K$ nearest neighbors in high-dimensional space, but not in low-dimensional space. 
 TW and CT assume values in $[0,1]$ and large values imply that neighborhoods are better preserved; $TW=1$ and $CT=1$ indicate perfect preservation of neighborhood ranking.

\subsection{Performance Comparison}
%

\begin{table}[tp]
		\centering
		\renewcommand{\arraystretch}{1.05}
		\caption{Supervised test-set performance}
		\label{tbl:fullysupervisedtestset}
		\scalebox{1}{ 
			\begin{tabular}{@{}ll|cc|cc@{}}
				\toprule
				\multicolumn{2}{c}{} & \multicolumn{2}{c}{\bf Q-LoS} & \multicolumn{2}{c}{\bf Q-NLoS} \\
				\midrule
				&  & FCNN & Siamese   & FCNN & Siamese  \\
				\midrule
				MDE  [m]   & & 7.31  & \textbf{6.69}         & 11.32   & \textbf{10.51}\\
				\midrule
				KS  &      &  0.146 & \textbf{0.107}      & 0.293 & \textbf{0.116}   \\	
				\midrule
				\multirow{3}{*}{TW} 
				& $K=1$   &  \textbf{0.988} & 0.987       & 0.966 & \textbf{0.968}   \\
				& $K=40$  &  \textbf{0.993} & \textbf{0.993}       & 0.978 & \textbf{0.982}  \\
				& $K=80$  &  0.994 & \textbf{0.995}       & 0.980 & \textbf{0.987} \\
				\midrule
				\multirow{3}{*}{CT} 
				& $K=1$   &  0.988 & \textbf{0.989}       & 0.965 & \textbf{0.966}   \\
				& $K=40$  &  0.993 & \textbf{0.994}       & 0.980 & \textbf{0.981} \\
				& $K=80$  &  \textbf{0.994} & \textbf{0.994}       & 0.983 & \textbf{0.984}   \\

				\bottomrule
			\end{tabular}
		}
\end{table}

\subsubsection{Supervised positioning}
\fref{tbl:fullysupervisedtestset} compares the test-set performance of the reference FCNN to that of the proposed Siamese network in the supervised scenario, i.e., ground truth location is available for all $N=2\,000$ UEs during training. We observe that the Siamese network outperforms the FCNN in almost all cases, except for the TW measure in the Q-LoS case with $K=1$ (where the TW is almost on par with the reference FCNN). The MDE and KS of the proposed Siamese network are both substantially smaller than that of the FCNN.
We note that, while both networks are trained in a conventional regression scenario, the Siamese networks achieve superior performance. Since $f_\bmtheta$ of the Siamese network is the same as the function of the FCNN, we conjecture that the improved performance is due to the fact that Siamese networks learn from $(N^2-N)/2$ data points (all distinct pairwise distances), which serves as a regularizer during SGD. In contrast, the FCNN learns its parameters  from only $N$ data points. 

\begin{table}[tp]
		\centering
		\renewcommand{\arraystretch}{1.05}
		\caption{Channel charting (unsupervised) training performance}
		\label{tbl:CCresultstrainig}
		\scalebox{1.0}{ 
			\begin{tabular}{@{}ll|ccc|ccc@{}}
				\toprule
				\multicolumn{2}{c}{} & \multicolumn{3}{c}{\bf Q-LoS} & \multicolumn{3}{c}{\bf Q-NLoS} \\
				\midrule
				&  &  Samm. & AE & Siam.    &  Samm. & AE & Siam.  \\
				\midrule
				KS &      &  0.999 & 0.980 &\textbf{ 0.951}     & 0.990 & 0.978 &  \textbf{0.939}\\
				\midrule	
				\multirow{3}{*}{TW} 
				& $K=1$   &  0.890 & \textbf{0.984 }&  0.874    & 0.969 & \textbf{0.988} &  0.965\\
				& $K=40$  &  0.865 & \textbf{0.951} & 0.870     & 0.972 & \textbf{0.986} & 0.967\\
				& $K=80$  &  0.851 & \textbf{0.933} &  0.864    & 0.972 & \textbf{0.985} & 0.967\\
				\midrule
				\multirow{3}{*}{CT} 
				& $K=1$   &  \textbf{0.986} & 0.983 & 0.984     & 0.980 & \textbf{0.984} & 0.981 \\
				& $K=40$  &  0.959 & \textbf{0.968} &  0.961    & 0.978 & \textbf{0.981} & 0.979\\
				& $K=80$  &  0.946 & \textbf{0.956} &  0.951    & 0.976 & \textbf{0.979} &  0.978\\

				\bottomrule
			\end{tabular}
		}
	\end{table}

\begin{table}[tp]
		\centering
		\renewcommand{\arraystretch}{1.05}
		\caption{Channel charting (unsupervised) test-set performance}
		\label{tbl:CCresultsvalidation}
		\scalebox{1.0}{ 
			\begin{tabular}{@{}ll|ccc|ccc@{}}
				\toprule
				\multicolumn{2}{c}{} & \multicolumn{3}{c}{\bf Q-LoS} & \multicolumn{3}{c}{\bf Q-NLoS} \\
				\midrule
				&  &  Samm. & AE & Siam.    &  Samm. & AE & Siam.  \\
				\midrule
				KS &      &  N/A & 0.985 & \textbf{0.957}     & N/A & 0.982 &  \textbf{0.942}\\				
				\midrule
				\multirow{3}{*}{TW} 
				& $K=1$   &  N/A & \textbf{0.940} & 0.843    & N/A & \textbf{0.982} &  0.966\\
				& $K=40$  &  N/A & \textbf{0.844} & 0.802     & N/A & \textbf{0.976} & 0.959\\
				& $K=80$  &  N/A & 0.735 &\textbf{0.763}    & N/A & \textbf{0.964} & 0.934\\
				\midrule
				\multirow{3}{*}{CT} 
				& $K=1$   &  N/A & 0.957 & \textbf{0.967}      & N/A & \textbf{0.980} & 0.976 \\
				& $K=40$  &  N/A & 0.862 & \textbf{0.885}    & N/A & \textbf{0.972} & 0.963\\
				& $K=80$  &  N/A & 0.749 & \textbf{0.786}    & N/A & \textbf{0.956} &  0.923\\
				\bottomrule
			\end{tabular}
		}
\end{table}

\subsubsection{Channel charting (unsupervised)}
Tables~\ref{tbl:CCresultstrainig} and \ref{tbl:CCresultsvalidation} show the training and test-set performance, respectively, of (unsupervised) CC. We also consider the training performance as conventional Sammon's mapping does not generate a parametric mapping---hence, results on the test set are not available.
Since CC is unable to perform absolute positioning, we omit MDE performance.
For the training performance, the AE consistently outperforms Sammon's mapping and the proposed Siamese network---the performance gap between the AE and Siamese network is small. The performance of the Siamese network and traditional Sammon's mapping is similar, indicating we were able to learn a parametric function for Sammon's mapping.
In terms of test-set performance, the Siamese network performs on par with the AE for the Q-LoS channel and only slightly worse for the Q-NLoS channel.

\begin{table}[tp]
		\centering
		\renewcommand{\arraystretch}{1.05}
		\caption{Semisupervised test-set performance}
		\label{tbl:semisupervised}
		\scalebox{1.0}{ 
			\begin{tabular}{@{}ll|cc|cc@{}}
				\toprule
				\multicolumn{2}{c}{} & \multicolumn{2}{c}{\bf Q-LoS} & \multicolumn{2}{c}{\bf Q-NLoS} \\
				\midrule
				&  & AE & Siamese   & AE & Siamese  \\
				\midrule
				MDE [m]  & & 13.35  & \textbf{10.62}         & 21.70   & \textbf{17.59}\\
				\midrule
				KS  &      &  0.324 &\textbf{0.275}       & 0.491 & \textbf{0.327}   \\			
				\midrule
				\multirow{3}{*}{TW} 
				& $K=1$   & \textbf{0.976} & \textbf{0.976} & \textbf{0.942} &0.932   \\
				& $K=40$  & 0.977 & \textbf{0.986}      & \textbf{0.961} & 0.953  \\
				& $K=80$  &  0.981 & \textbf{0.988}      &\textbf{0.965} & 0.961 \\
				\midrule
				\multirow{3}{*}{CT} 
				& $K=1$   &  0.979 & \textbf{0.980}       & 0.953 & \textbf{0.954}   \\
				& $K=40$  &  0.979 & \textbf{0.987}       & 0.954 & \textbf{0.958} \\
				& $K=80$  &  0.982 & \textbf{0.989}       & 0.957 & \textbf{0.965}   \\
				\bottomrule
			\end{tabular}
		}
\end{table}

\subsubsection{Semisupervised positioning}
\fref{tbl:semisupervised} shows the validation performance of semi-supervised positioning, where we used the representation-constrained autoencoder (AE) as in~\cite{huang2019improving} for semisupervised positioning. We assumed that only 10\% of the UE locations were available during training. 
We see that the Siamese network and the representation-constrained AE perform comparably well for both channel scenarios.

 \begin{table}[tp]
		\centering
		\renewcommand{\arraystretch}{1.05}
		\caption{Supervised test-set performance with~only~10\%~training~data}
		\label{tbl:semisupervisedcomparison}
		\scalebox{1.0}{ 
			\begin{tabular}{@{}ll|cc|cc@{}}
				\toprule
				\multicolumn{2}{c}{} & \multicolumn{2}{c}{\bf Q-LoS} & \multicolumn{2}{c}{\bf Q-NLoS} \\
				\midrule
				&  & FCNN & Siamese   & FCNN & Siamese  \\
				\midrule
				MDE  [m]   & & 13.33  & \textbf{12.81}         & 17.64   & \textbf{17.33}\\
				\midrule
				KS  &      &  0.401 & \textbf{0.154}      & 0.395 & \textbf{0.288}   \\
				\midrule
				\multirow{3}{*}{TW} 
				& $K=1$   &  \textbf{0.981} & 0.965     & \textbf{0.938} & {0.929}   \\
				& $K=40$  &  \textbf{0.986} & {0.976}       & 0.953 & \textbf{0.954}  \\
				& $K=80$  &  \textbf{0.988} & {0.981}  & \textbf{0.958} & \textbf{0.958} \\
				\midrule
				\multirow{3}{*}{CT} 
				& $K=1$   &  \textbf{0.983} & {0.977}       & 0.947 & \textbf{0.949}   \\
				& $K=40$  &  \textbf{0.986} & {0.976}       & \textbf{0.956} & \textbf{0.956} \\
				& $K=80$  &  \textbf{0.988} & {0.979}       & 0.961 & \textbf{0.962}   \\

				\bottomrule
			\end{tabular}
		}
\end{table}

\begin{table}[tp]
	\centering
	\renewcommand{\arraystretch}{1.05}
	\caption{Semisupervised T-intersection test-set performance}
	\label{tbl:tintersection}
	\scalebox{1.0}{ 
		\begin{tabular}{@{}ll|c|c@{}}
			\toprule
			\multicolumn{2}{c}{} & \multicolumn{1}{c}{\bf Q-LoS} & \multicolumn{1}{c}{\bf Q-NLoS} \\
			\midrule
			MDE [m]  &   & {2.95}            & {5.32}\\
			\midrule
			KS  &       & {0.188}        & {0.471}   \\
			\midrule
			\multirow{3}{*}{TW} 
			& $K=1$    & {0.560}        & 0.560   \\
			& $K=40$  & {0.605}        & 0.579  \\
			& $K=80$   & {0.755}        & 0.729 \\
			\midrule
			\multirow{3}{*}{CT} 
			& $K=1$    & {0.970}        & 0.947    \\
			& $K=40$   & {0.951}        & 0.934  \\
			& $K=80$   & {0.974}        & 0.944   \\
			\bottomrule
		\end{tabular}
	}
\end{table}

In order to determine whether semisupervised positioning has any advantage over fully-supervised positioning with the same number of observed UE locations, we train a FCNN and a Siamese network with only 10\% known UE locations ($N=200$). Note that this is the same number of  known anchor positions as for the semisupervised experiment in \fref{tbl:semisupervised}; in the semisupervised case, however, we use all of the $N=2\,000$ channel features during training (but only $N=200$ known UE locations). 
As we can see by comparing the performance of Siamese networks in \fref{tbl:semisupervisedcomparison} with \fref{tbl:semisupervised} for the Q-LoS channel model, using only 10\% of the data in the supervised setting results in inferior performance in terms of MDE, KS, TW, and CT, compared to the semisupervised case.
Quite surprisingly, for the more challenging Q-NLoS channel model, semisupervised learning did not show any advantage over training from only 10\% data. 
This implies that the proposed method must be improved even further if one wants to take advantage of semisupervised learning. 
 
\begin{figure*}[tp]
\centering
\begin{subfigure}{0.325\textwidth}
\centering
\includegraphics[width=0.95\textwidth]{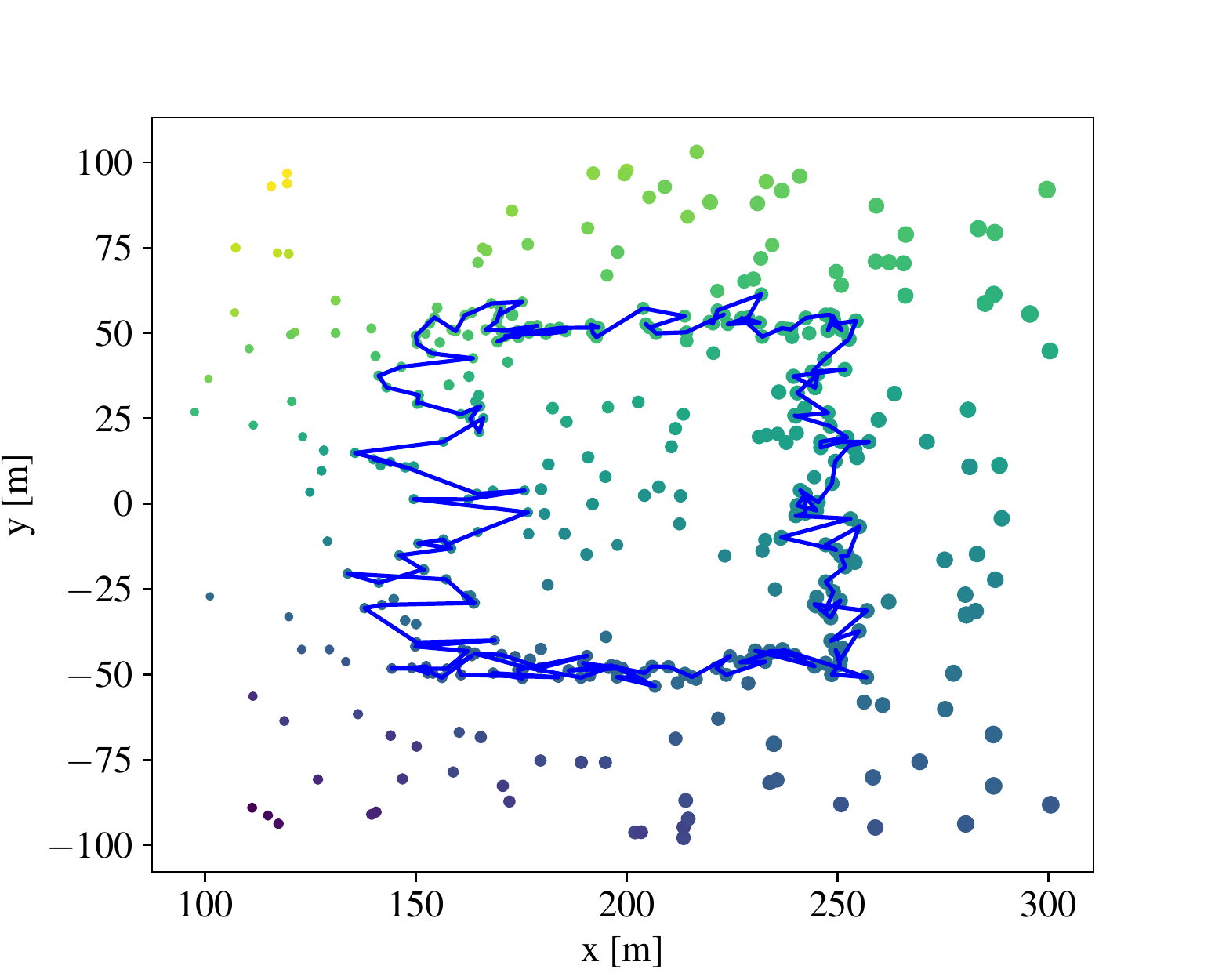}
\caption{Supervised;  $\textit{MDE} = 6.69$, $\textit{KS} = 0.107$, $\textit{TW} = 0.995$, and $\textit{CT} = 0.994$.}
\end{subfigure}
\hfill
\begin{subfigure}{0.325\textwidth}
\centering
\includegraphics[width=0.95\linewidth]{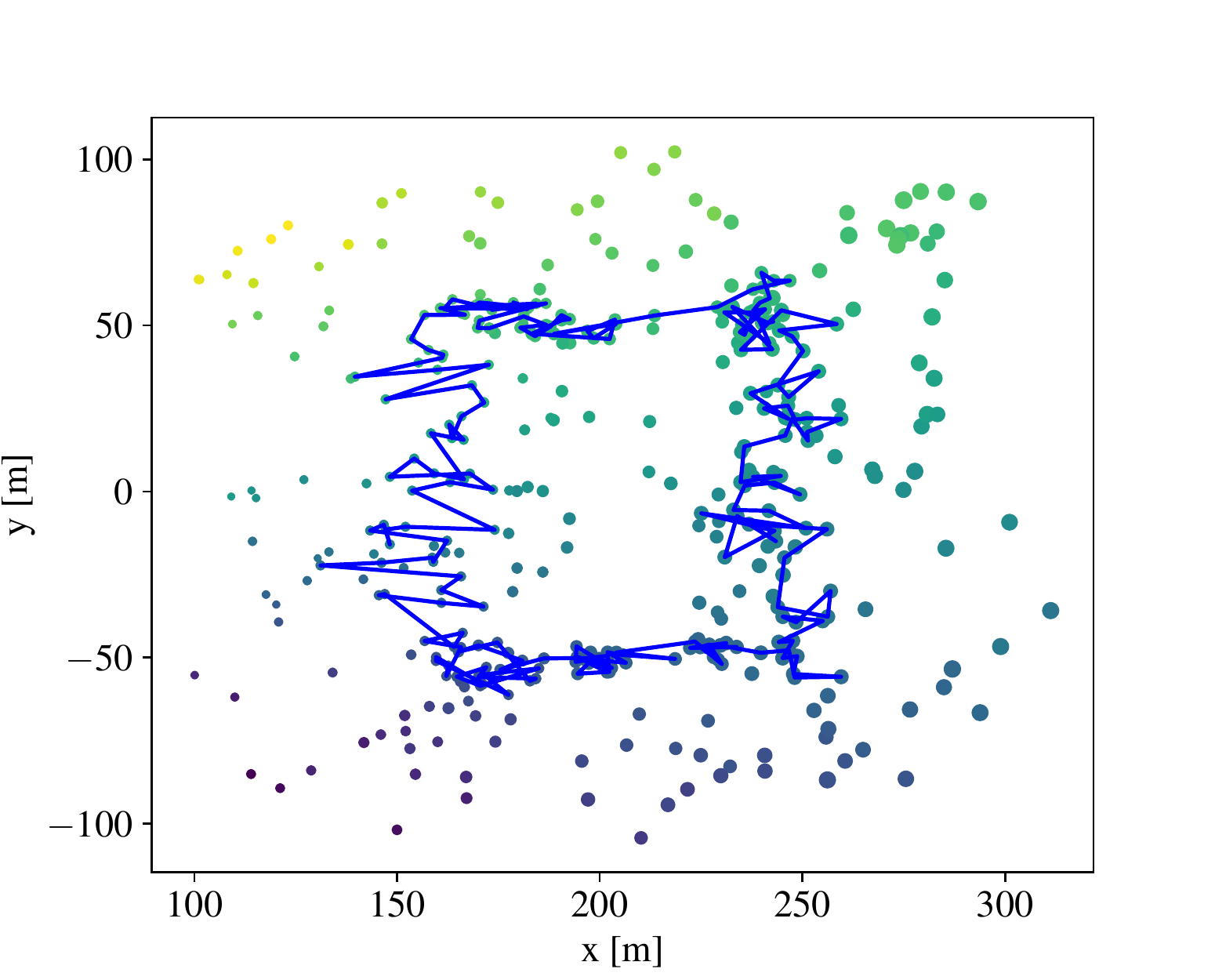}
\caption{Semisupervised; $\textit{MDE} =  10.62$, $\textit{KS} =  0.275$, $\textit{TW} =  0.988$, and $\textit{CT} = 0.989$.}
\end{subfigure}
\hfill
\begin{subfigure}{0.325\textwidth}
\centering
\includegraphics[width=0.95\linewidth]{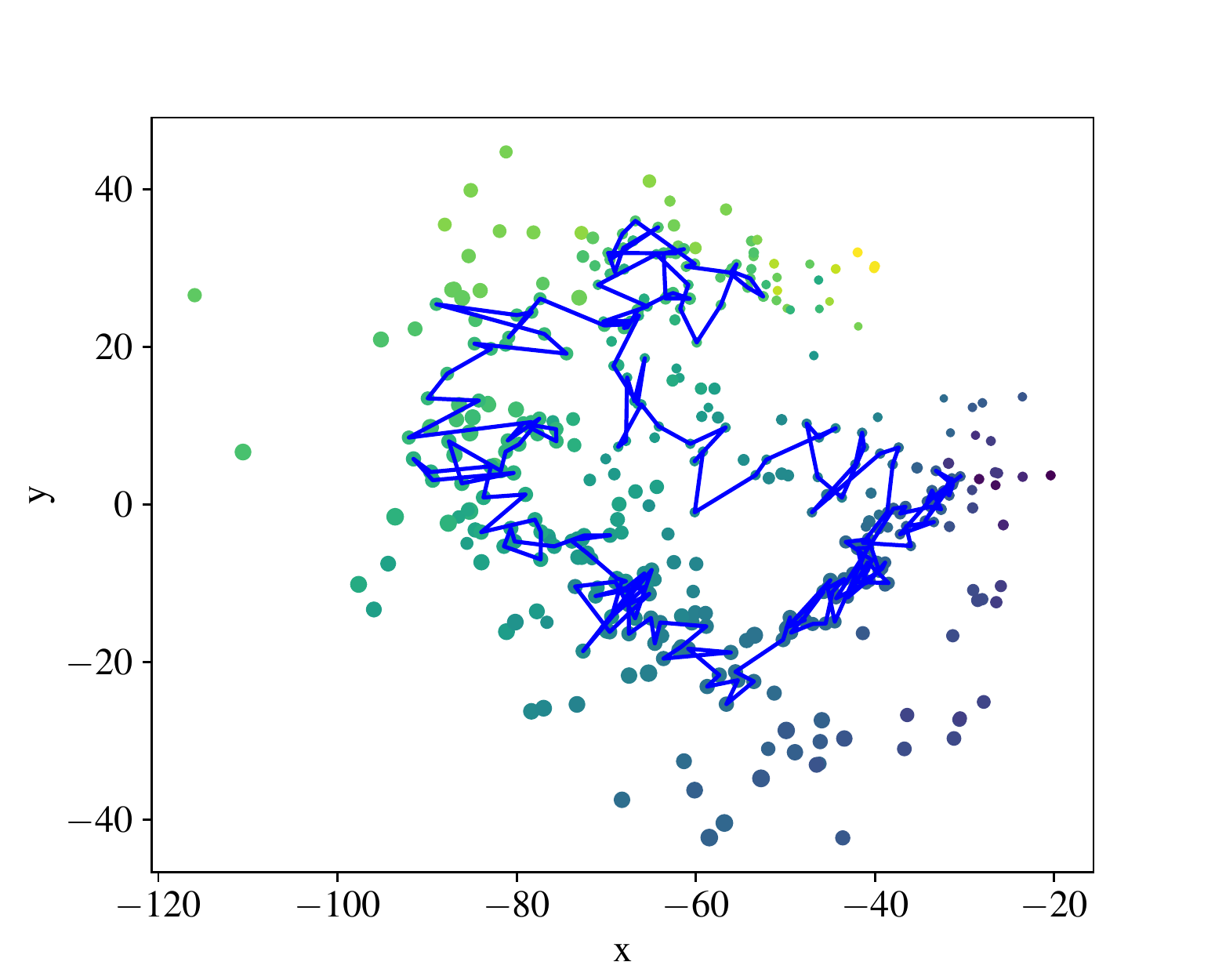}
\caption{Channel charting (unsupervised);  \textit{KS} = $0.957$, \textit{TW} = $0.763$, and \textit{CT} = $0.786$.}
\end{subfigure}
\caption{Visualization of predicted locations via Siamese networks for a Q-LoS channel. \textit{TW} and \textit{CT} values are evaluated using $K = 80$ nearest neighbors.}
\label{fig:visLoS}
\end{figure*}

\begin{figure*}[tp]
\centering
\begin{subfigure}{0.325\textwidth}
\centering
\includegraphics[width=0.95\textwidth]{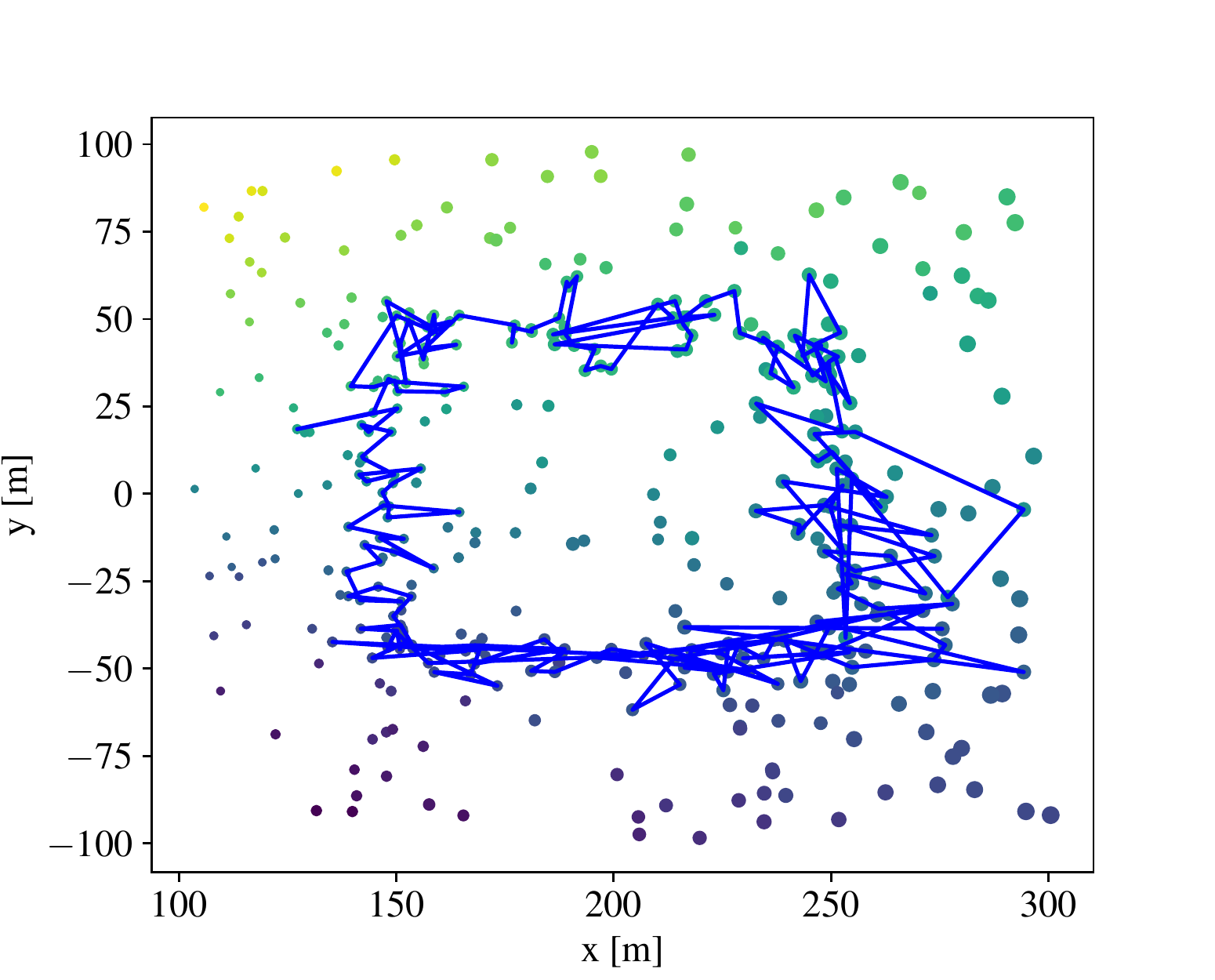}
\caption{Supervised; $\textit{MDE} =  10.51$, $\textit{KS} =  0.116$, $\textit{TW} =  0.987$, and $\textit{CT} =  0.984$.}
\end{subfigure}
\hfill
\begin{subfigure}{0.325\textwidth}
\centering
\includegraphics[width=0.95\linewidth]{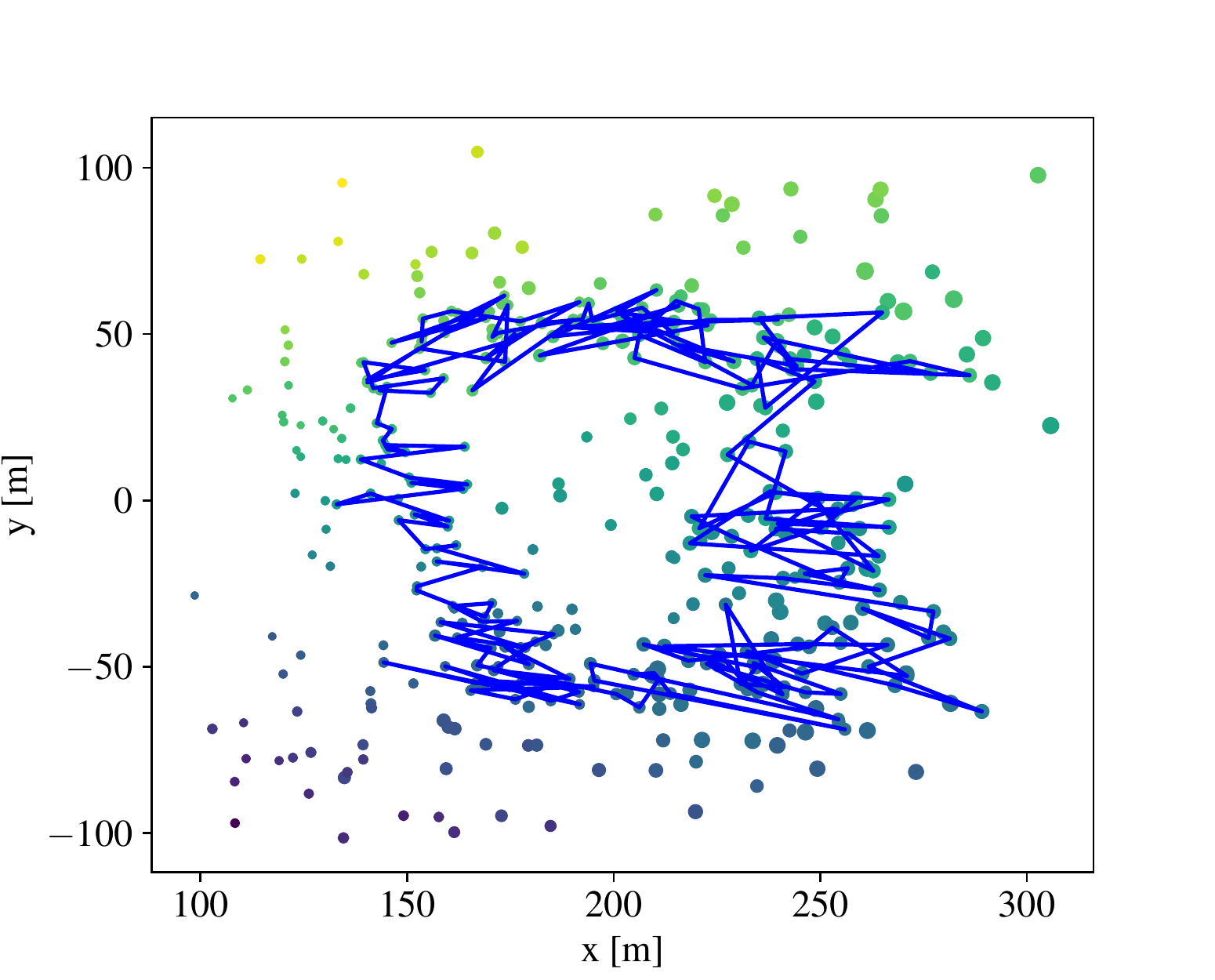}
\caption{Semisupervised; $\textit{MDE} =  17.59$, $\textit{KS} =  0.327$, $\textit{TW} =  0.961$, and $\textit{CT} =  0.965$.}
\end{subfigure}
\hfill
\begin{subfigure}{0.325\textwidth}
\centering
\includegraphics[width=0.95\linewidth]{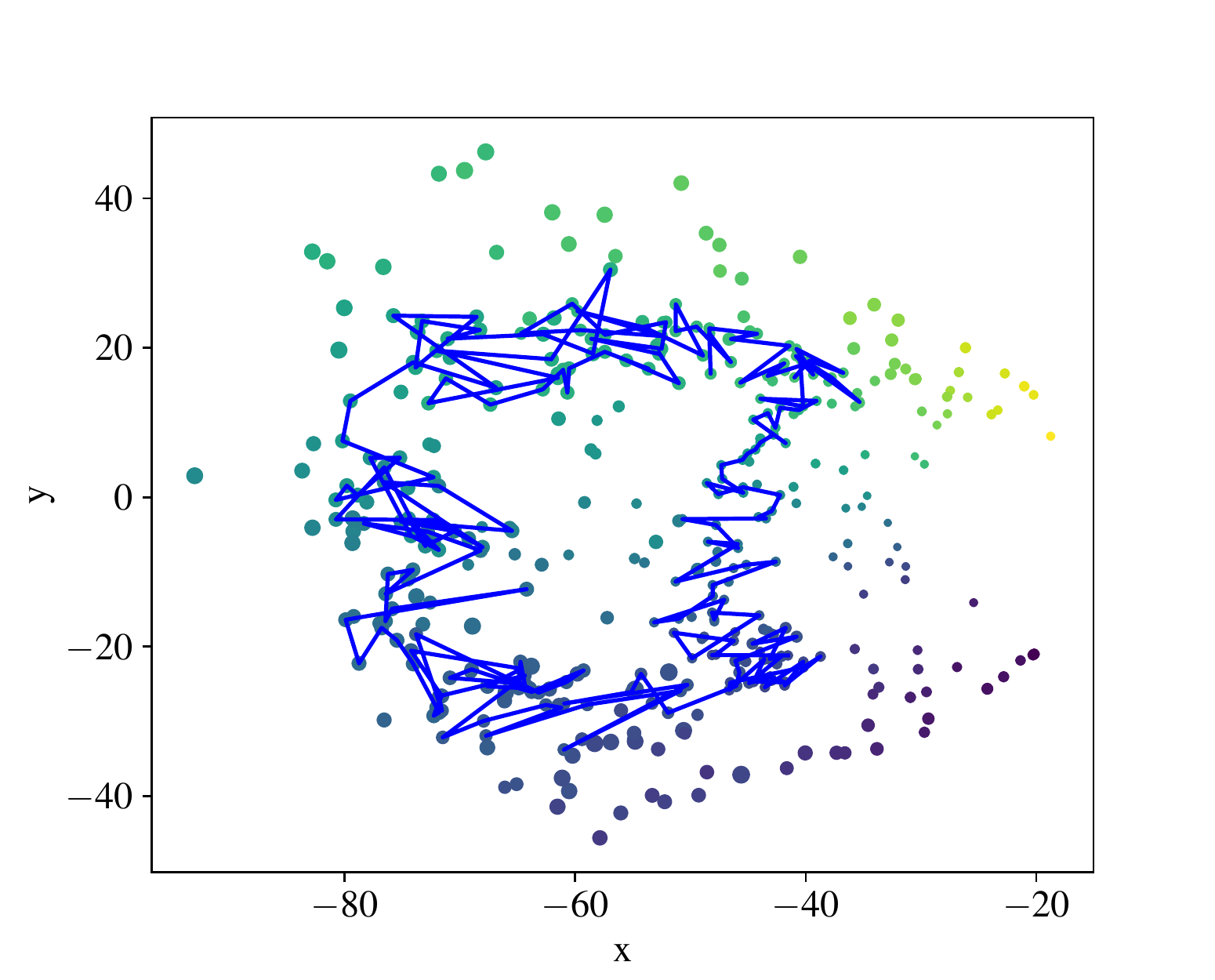}
\caption{Channel charting (unsupervised); $\textit{KS} = 0.942$, $\textit{TW} =  0.934$, and $\textit{CT} =  0.923$.}
\end{subfigure}
\caption{Visualization of predicted locations via Siamese networks for a Q-NLoS channel. \textit{TW} and \textit{CT} values are evaluated using $K = 80$ nearest neighbors.}
\label{fig:visNLoS}
\end{figure*}

\subsubsection{Visualization of Positioning and CC Results}
Figures \ref{fig:visLoS} and \ref{fig:visNLoS} show the learned positions on the test set for Q-LoS and Q-NLoS channel with Siamese networks, respectively. Note that the same unified Siamese network architecture was used to obtain all of these results. 
As we can see, CSI-based positioning in the supervised setting works well on both LoS and non-LoS channels, whereas the results for the non-LoS case are, as expected, slightly less accurate. The learned positions for the semisupervised setting are only slightly worse than those for the fully supervised scenario. 
The (unsupervised) CC results  do not allow absolute positioning, but one can clearly see that local geometry as well as global geometry is well preserved.

\begin{figure*}[tp]
\centering
\begin{subfigure}{0.325\textwidth}
\centering
\includegraphics[width=0.95\textwidth]{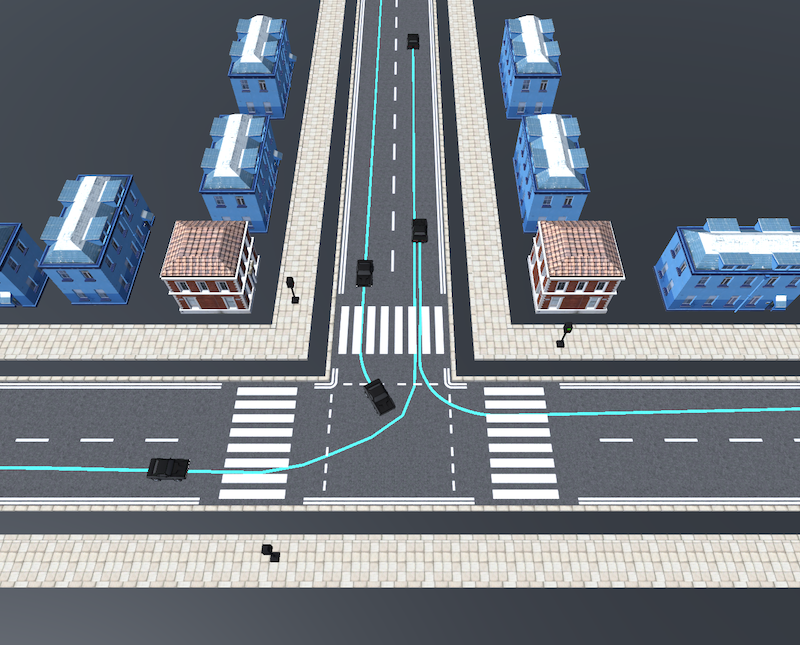}
\caption{Screenshot of T-intersection simulator.}
\end{subfigure}
\hfill
\begin{subfigure}{0.325\textwidth}
\centering
\includegraphics[width=0.95\linewidth]{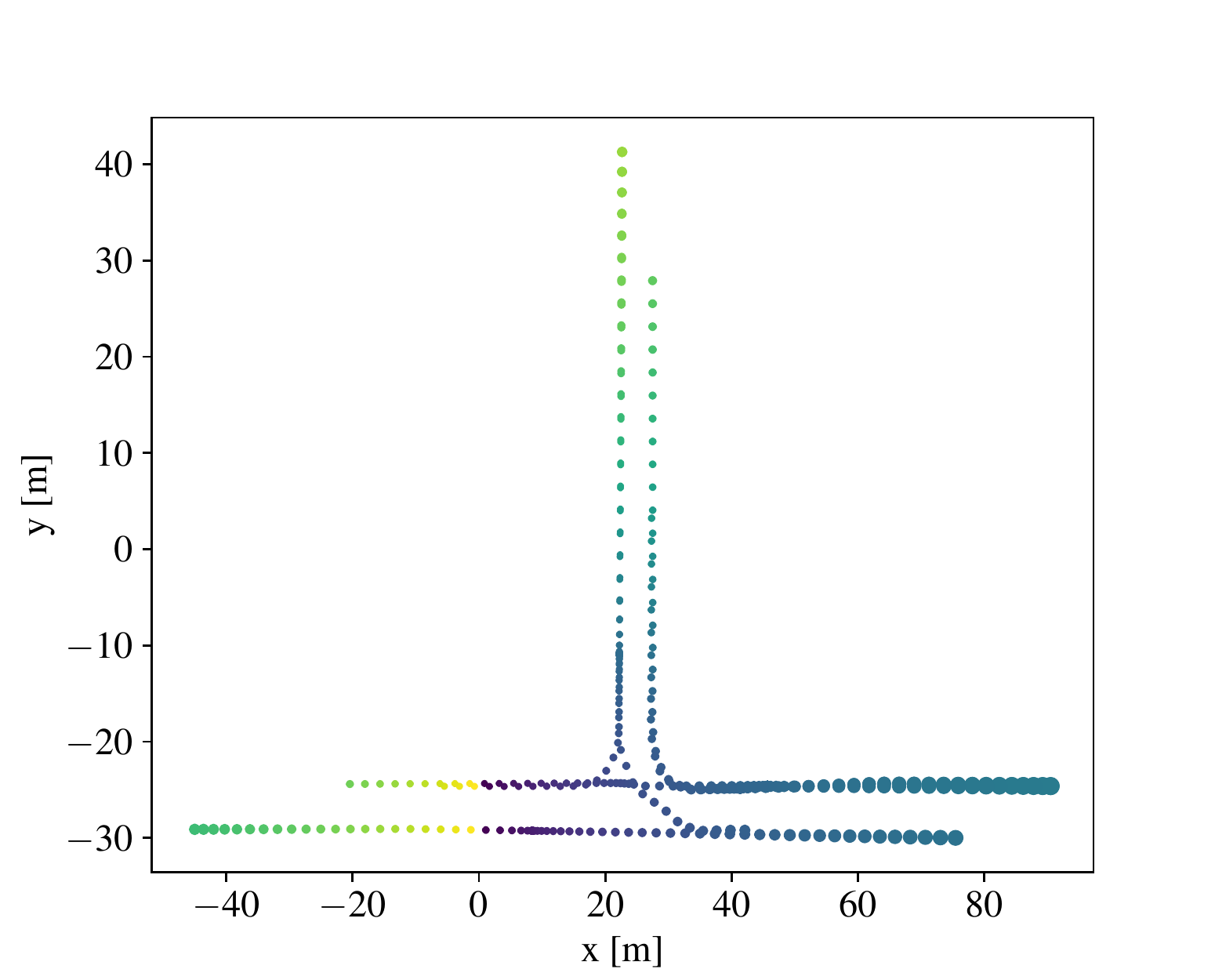}
\caption{Ground truth test-set positions.}
\end{subfigure}
\hfill
\begin{subfigure}{0.325\textwidth}
\centering
\includegraphics[width=0.95\linewidth]{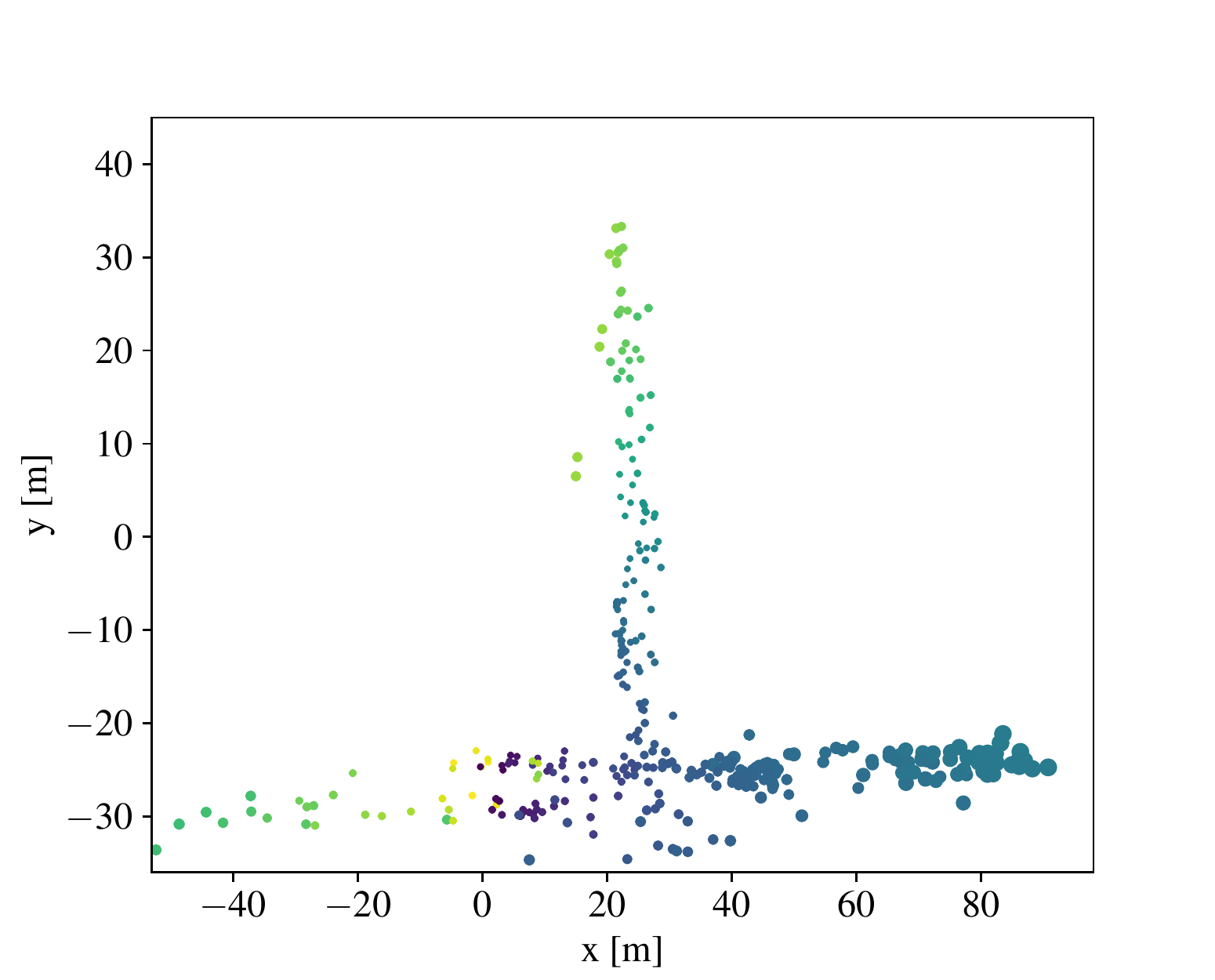}
\caption{Predicted locations via Siamese networks.}
\end{subfigure}
\caption{Visualization of predicted locations using a Siamese network on UE positions sampled from a T-intersection traffic simulator using a Q-LoS channel model. Even from only 20 measured traces, new UEs can be traced accurately. The performance of predicted locations is $\textit{MDE}=2.95$, $\textit{KS}=0.188$, $\textit{TW}=0.755$, and $\textit{CT}=0.974$, where the TW and CT values are evaluated using $\textit{K}=80$ nearest neighbors.}
\label{fig:tintersection}
\vspace{-0.2cm}
\end{figure*}

\subsection{T-Intersection Scenario}
We now show a more realistic scenario in which cars are passing through a T-intersection from three directions as depicted in \fref{fig:tintersection}(a). The car movement was simulated using the Unity game engine. 
Each car creates what we call a ``trace'' that consists of CSI associated to its position collected over time. Example traces of UE position are shown in \fref{fig:tintersection}(b). 
We trained a Siamese network from the CSI obtained by observing $20$ such traces and then use the resulting network for positioning of new, unseen UE locations on the same T-intersection. 
\fref{fig:tintersection}(c) shows the learned UE positions for a test set consisting of 20 new traces. Clearly, the Siamese network is able to accurately place new UE positions in real space. 
\fref{tbl:tintersection} shows the associated performance measures. We see that an MDE of only $2.95$\,m and $5.32$\,m is attainable for the Q-LoS and Q-NLoS channel, respectively, which demonstrates that Siamese networks are able to accurately position UEs from very small datasets generated by realistic motion over time.

\section{Conclusions}
We have shown that Siamese networks can be used to implement parametric Sammon's mapping, which enables parametric channel charting (CC) from channel-state information (CSI). In addition, we have shown that the same neural network architecture can be used to include partially annotated data of user equipment (UE) positions, which enables supervised as well as semisupervised positioning. 
By comparing the proposed Siamese network to that of fully-connected neural networks, autoencoders, and traditional Sammon's mapping, we have demonstrated that our approach performs on par with or superior to baseline methods, but with a unified neural network architecture. 
Moreover, we have demonstrated that, for Siamese networks, semisupervised training is able to outperform supervised training under line-of-sight channel conditions---this implies that including unlabeled CSI measurements for positioning can be beneficial. 
Finally, we have shown that Siamese networks are able to perform accurate positioning for scenarios with realistic UE motion, even for very small datasets.

%
%
There are many opportunities for extensions of this work.
An evaluation of Siamese networks with real-world CSI is already ongoing.
An extension of Siamese networks to situations with simultaneous connection to multiple BSs, as put forward in \cite{ponnada2019out,deng2018multipoint} for CC, is left for the future. 
Finally, the development of methods that further improve semisupervised training for challenging propagation scenarios is an open research problem.

	\balance
	\bibliographystyle{IEEEtran}
	\bibliography{VIPabbrv,confs-jrnls,iclr2019_conference}
	
	\balance
	
\end{document}